\documentclass{article}

\usepackage[preprint]{neurips_2024}

\usepackage[utf8]{inputenc} 
\usepackage[T1]{fontenc}    
\usepackage{hyperref}       
\usepackage{url}            
\usepackage{booktabs}       
\usepackage{amsfonts}       
\usepackage{nicefrac}       
\usepackage{microtype}      
\usepackage{xcolor}         

\usepackage{pdfpages}
\usepackage{amsmath}     
\usepackage{amssymb}     
\usepackage{bm}          
\usepackage{mathtools}   

\usepackage{algorithm}
\usepackage{algorithmicx}
\usepackage{algpseudocode}

\usepackage{graphicx}    
\usepackage{booktabs}    
\usepackage{siunitx}     
\usepackage{hyperref}    

\usepackage{graphicx}
\usepackage{amsmath}
\usepackage{multirow}
\usepackage{comment}
\usepackage{float}
\usepackage[font=small]{caption}

\usepackage[table]{xcolor}
\usepackage{booktabs}
\usepackage{array}
\definecolor{GenotypeBlue}{HTML}{D0E4F5}
\definecolor{ExpressionPurple}{HTML}{E6E0F8}
\definecolor{MetaboliteRed}{HTML}{F8D7DA}

\title{Biology-informed neural networks learn nonlinear representations from omics data to improve genomic prediction and interpretability}

\author{%
  Katiana Kontolati\thanks{Corresponding authors} \\
  Bayer Crop Science\\
  \texttt{katiana.kontolati@bayer.com} \\
  \And
  Rini Jasmine Gladstone\\
  Bayer Crop Science\\
  \texttt{rinijasmine.gladstone@bayer.com} \\
  \And
  Ian W. Davis \\
  Bayer Crop Science\\
  \texttt{ian.davis@bayer.com} \\
  \And
  Ethan Pickering$^{\ast}$ \\
  Bayer Crop Science\\
  \texttt{ethan.pickering@bayer.com} \\
}

\begin{document}

\maketitle

\begin{abstract}

We extend biologically-informed neural networks (BINNs) for genomic prediction (GP) and selection (GS) in crops by integrating thousands of single-nucleotide polymorphisms (SNPs) with multi-omics measurements and prior biological knowledge. Traditional genotype-to-phenotype (G2P) models depend heavily on direct mappings that achieve only modest accuracy, forcing breeders to conduct large, costly field trials to maintain or marginally improve genetic gain. Models that incorporate intermediate molecular phenotypes such as gene expression can achieve higher predictive fit, but they remain impractical for GS since such data are unavailable at deployment or design time. BINNs overcome this limitation by encoding pathway-level inductive biases and leveraging multi-omics data only during training, while using genotype data alone during inference. By embedding omics-derived priors directly into the network architecture, BINN outperforms conventional models in low-data ($n < p$) regimes and enables sensitivity analyses that expose biologically meaningful traits. Applied to maize gene-expression and multi-environment field-trial data, BINN improves rank-correlation accuracy by up to 56\% within and across subpopulations under sparse-data conditions and nonlinearly identifies genes that GWAS/TWAS fail to uncover. With complete domain knowledge for a synthetic metabolomics benchmark, BINN reduces prediction error by 75\% relative to conventional neural nets and correctly identifies the most important nonlinear pathway. Importantly, both cases show  highly sensitive BINN latent variables correlate with the experimental quantities they represent, despite not being trained on them.  This suggests BINNs learns biologically-relevant representations, nonlinear or linear, from genotype to phenotype. Together, BINNs establish a framework that leverages intermediate domain information to improve genomic prediction accuracy and reveal nonlinear biological relationships that can guide genomic selection, candidate gene selection, pathway enrichment, and gene-editing prioritization.

\end{abstract}






\section{Introduction}

\begin{figure}[ht]
  \centering
  \includegraphics[width=0.75\linewidth]{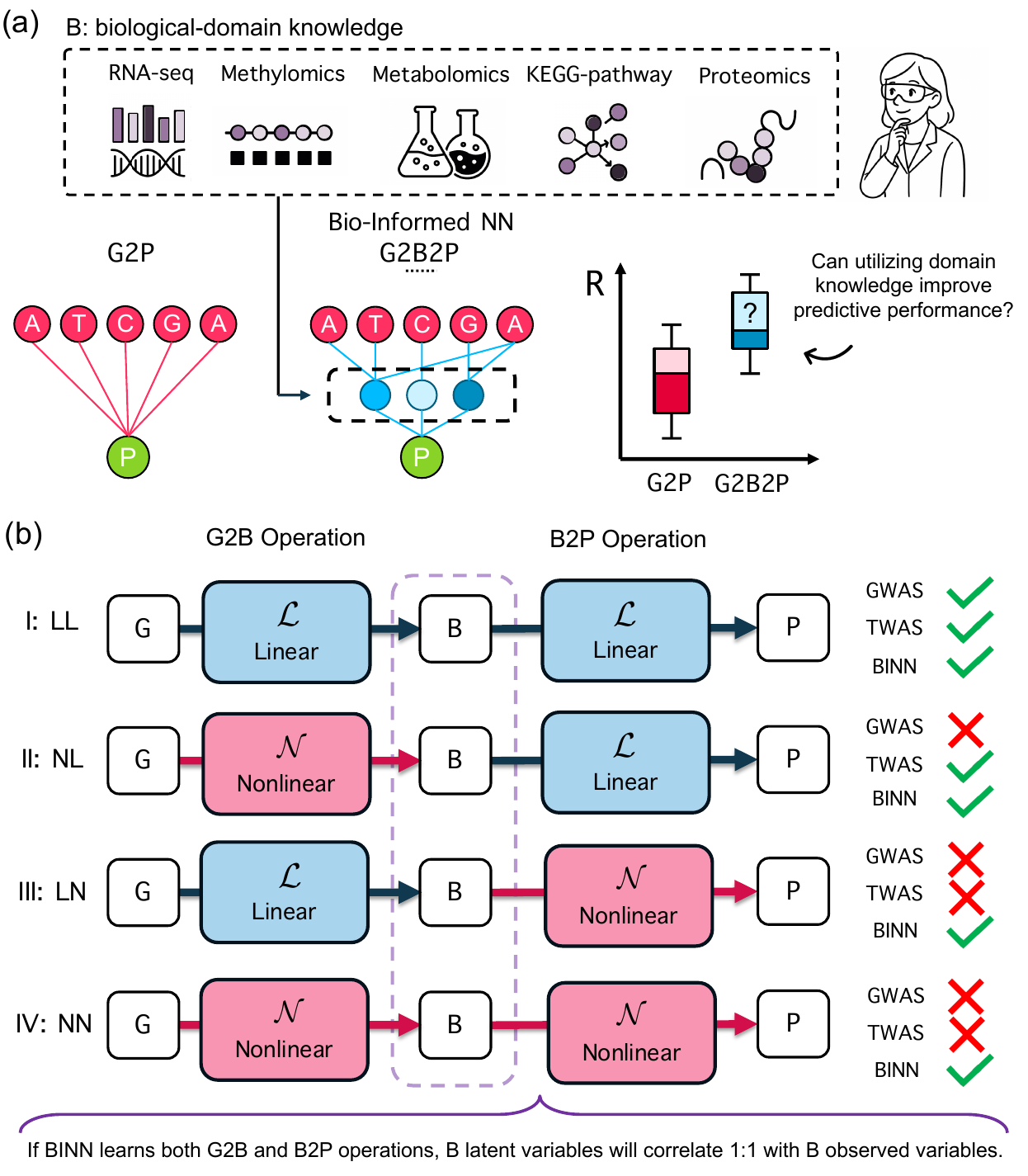}
  \caption{\textbf{Biology-informed neural network framework embed domain knowledge for enhanced genomic prediction and learning nonlinear biological relationships.} $a)$ Conventional G2P models use genotype alone, leaving rich functional knowledge underutilized. BINNs embed curated biology (from RNA-seq (expression), methylomics (DNA methylation), metabolomics, KEGG pathway annotations, and proteomics) directly into the architecture in the form of pathway structure, regulatory priors, and sparsity constraints, to boost predictive accuracy while preserving practical utility. $b)$ Four representative cases where GWAS, TWAS, and BINN are valid for analyzing genomic, transcriptomic, and phenomic datasets. Only BINN permits association under general nonlinearity (with the assumption the trained BINN model is accurate).}
  \label{fig:BINN-framework}
\end{figure}

Feeding 10 billion people by 2050 will require faster, more reliable crop improvement. In both conventional breeding—crossing and selecting offspring—and genome editing—making targeted edits and selecting edited lines—progress hinges on accurate genotype-to-phenotype (G2P) prediction to enable genomic selection (GS) \citep{meuwissen2001prediction,crossa2017genomic}. GS improves crop improvement efficiency by permitting early selection at the seed or edit stage and reducing dependence on slow, costly, multi-location field trials \citep{heffner2009genomic,varshney2021genomic}. With a well-calibrated model trained on prior genotype–phenotype data, a single low-cost genotyping assay can substantially reduce phenotyping requirements in both space and time \citep{crossa2017genomic}. A wide range of models has been applied to GS—including linear mixed models; machine-learning methods such as random forests and kernel approaches; and deep neural networks—with performance varying by trait architecture, data volume, and environment \citep{breiman2001random,yegnanarayana2009artificial}. Current approaches fall short of capturing the true biological processes from genotype to phenotype, and the field has a long way to go toward accurate, mechanistically grounded prediction. 


Despite the surge of countless AI models and architectures, linear mixed modeling approaches remain state-of-the-art in G2P studies, with no other class of model consistently outperforming them across crops, populations, traits, years, and locations~\cite{alemu2024genomic,azodi2019benchmarking,washburn2025global}.  Although there are several adaptations to linear models, most are quite similar.  For instance, ridge regression~\cite{mcdonald2009ridge} and its adaptation using best linear unbiased predictor, rrBLUP~\cite{endelman2011ridge}, trained using genotypic (marker) data are identical, except rrBLUP chooses the regularization parameter, $\alpha$, as a ratio of variances instead of by cross-validation~\cite{ogutu2012genomic}.  Likewise a genomic BLUP or GBLUP~\cite{clark2013genomic} is mathematically equivalent to rrBLUP, but with compute efficiency gains when the number of lines is less than the number of markers~\cite{jacquin2016unified}, while the "Bayesian alphabet" models (BayesA, BayesB, BayesC, etc)~\cite{gianola2009additive} differ in allowing different markers to have different priors~\cite{montesinos2022bayesian}. 
And although several promising nonlinear artificial neural network (ANN) architectures have been proposed in the literature~\cite{gao2023soydngp, ma2017deepgs, wang2025cropformer}, deep learning methods have yet to show consistent improvements over traditional models and have not been adopted at scale in breeding pipelines \cite{montesinos2021review}. This is partly due to over-parameterization and the lack of tailored deep learning architectures. Further gains are unlikely without injecting biological structure such as pathways, interactions, and regulatory programs into the model, a role for which flexible AI architectures are well suited. 






A natural route to inject such biological structure is to integrate intermediate multi-omics signals—transcripts, proteins, metabolites—as scaffolds between genotype and phenotype, enriching G2P models with mechanistic constraints~\cite{atwell2010genome,azodi2020transcriptome,christensen2021genetic,heffner2009genomic, korte2013advantages,meuwissen2001prediction,  riedelsheimer2012genomic}. Approaches such as multi-staged analysis~\cite{ritchie2015methods} and transcriptome-wide association studies (TWAS)~\cite{gamazon2015gene, wainberg2019opportunities} are proposed to integrate omics data, such as gene expression levels, between genotype and phenotype for association studies. Traditional linear mixed models, while capable of including additional covariates, are fundamentally limited in how they integrate multi-omics information. In practice, they treat each data modality as a flat feature set and cannot impose the hierarchical, pathway-level constraints that capture the flow from genotype through intermediate traits to phenotype. This lack of architectural flexibility prevents them from leveraging richer, layered representations, such as gene‐to‐metabolite cascades or expression-driven subnetworks, that can substantially boost predictive power. In contrast, ANNs provide the flexibility to integrate multiple high-dimensional data streams such as genomics, transcriptomics, proteomics, lipidomics, etc., for tasks ranging from regression and classification to unsupervised representation learning, a capability that traditional linear models lack \cite{ma2022omicsgcn, nguyen2020deepprog, wang2021mogonet, wang2023dnngp, yang2021mvae}. In practice, downstream omics are not available for novel genotypes at design or deployment, so they can only inform training—via priors, pretraining, or architectural constraints—rather than serve as inference-time features. 
Biologically-informed neural networks (BINNs) have begun to provide model flexibility in embedding omics data and domain-specific knowledge such as gene, protein, or pathway relationships, directly into their architecture and training objectives to improve predictive accuracy, interpretability, and extrapolative potential. Analogous to physics-informed neural networks (PINNs) that incorporate governing equations into their loss functions~\cite{raissi2019physics}, BINNs impose biologically plausible sparsity patterns on connections, reducing parameter count and data requirements while enabling efficient integration of heterogeneous multi-omics inputs. This inductive bias not only curbs overfitting but also aligns latent representations with known biology. BINNs have been applied across population genomics and biomedicine, including cancer subtype prediction, drug response modeling, and survival analysis in frameworks like GenNet~\cite{van2021gennet}, proteomics biomarker discovery~\cite{hartman2023interpreting}, hierarchical pathway networks for prostate cancer~\cite{elmarakeby2021biologically}, and multi-omics inferral of smoking status, age, and LDL from the BIOS cohort~\cite{van2024phenotype}, demonstrating consistent gains in performance and stability. 
Despite their promise, existing BINNs enforce a rigid, fully interpretable mapping of neurons to biological entities, limiting architectural flexibility and nonlinear modeling, and in most cases still depend on omics measurements at deployment, undermining their practicality in plants and crop design. Furthermore, although existing BINN approaches employ ANN-based models, their method for ranking biological entities such as genes typically relies on assessing marginal associations through learned single-edge weights, an approach conceptually similar to traditional genome-wide association studies (GWAS) or TWAS. Consequently, these models tend to highlight already known significant genes while overlooking those involved in nonlinear or epistatic interactions.

Here we design a BINN architecture to align with genomic selection and design of crops, explicitly integrating omics data as intermediate variables, removing the need for omics data at test/design while still allowing for biological interpretability and targeted pathway prioritization (Figure \ref{fig:BINN-framework}). Our key contributions are summarized below:  
\setlength{\itemsep}{0pt}
\begin{enumerate}
    \item A BINN architecture balancing tunable sparsity, layered nonlinearity and practicality. 
    \item Superior sparse-data performance with up to 56\% test-set rank correlation increase and 75\% reduction in predictive error over G2P baselines, across and within populations.
    \item Demonstrated ability to transfer across related populations, maintaining strong predictive accuracy when genetic background is partially shared.
    \item A novel BINN-derived sensitivity analysis framework that associates biologically meaningful intermediate traits to phenotype that are beyond the reach of conventional GWAS and TWAS approaches.
  \item A scalable, practical framework for plant and crop genomic selection and design that marries multi-omics-driven training with genotype-only deployment. 
\end{enumerate}

\section{Results} 





We develop and evaluate BINN models informed by two distinct types of intermediate omics—\textbf{transcriptomics} and \textbf{metabolomics}—to predict phenotypes directly from genotype. Our first case study leverages transcriptomic profiles measured in a large-scale maize field experiment\citep{torres2024population}, where lines from multiple heterotic groups were grown under real agronomic conditions. This setting provides a realistic testbed for BINNs: gene expression captures environmentally modulated regulatory activity, offering a richer signal than raw marker data. In this context, we observe promising gains over conventional G2P models, but also important caveats due to the noise, heterogeneity, and incomplete pathway knowledge that naturally accompany field-derived omics measurements. In contrast, our second case study is a synthetic benchmark based on metabolomic traits simulated through an ordinary differential equation (ODE) model of shoot branching\citep{bertheloot2020sugar}. Unlike the maize field data, this framework provides “perfect” domain knowledge of the true causal pathways, enabling us to rigorously stress-test BINNs and evaluate their ability to recover mechanistic structure when the ground truth is fully known. Together, these complementary case studies - one grounded in realistic breeding data, the other in controlled synthetic biology - allow us to assess both the practical utility and the methodological limits of BINNs. Table~\ref{tab:case-studies} summarizes implementation details; full configurations and preprocessing steps can be found in the Supplementary Section.

\begin{table}[h]
\centering
\footnotesize
\caption{Implementation components for the maize TWAS dataset and synthetic shoot-branching case studies.}
\label{tab:case-studies}
\begin{tabular}{@{} l  l  l @{}}
\toprule
\textbf{Attribute}                      & \textbf{Maize TWAS Dataset}              & \textbf{Synthetic Shoot-Branching}                       \\
\midrule
Input SNPs/Genes                        & \(\sim\)20,000 SNPs per trait                   & 1,600 genes                                              \\
Intermediate Omics Data                 & Transcriptomics                          & Metabolomics                                            \\ 
\# Intermediate Traits                  & \(\sim\)1,000 genes per trait   & 4 (auxin, sucrose, cytokinin, strigolactone)            \\
\# Output Phenotypes                    & 4 (anthesis NE, MI; silking NE, MI)      & 1 (time to bud outgrowth)                                       \\
Trait Selection Method                  & ElasticNet regression                         & From known ODE model                                    \\
SNP Mask Construction                   & eQTL mapping of selected genes     & ODE‐derived gene-metabolite mappings                    \\
Loss Function                           & MSE                                         & Soft-constrained MSE                                     \\
Reference                               & Torres-Rodriguez et al.~\cite{torres2024population}      & Bertheloot et al.~\cite{bertheloot2020sugar} \& Powell et al.~\cite{powell2022investigations} \\
\bottomrule
\end{tabular}
\end{table}

\subsection{Transcriptomics-derived BINN} 

\begin{figure}[ht!]
  \centering
\includegraphics[width=0.9\linewidth]{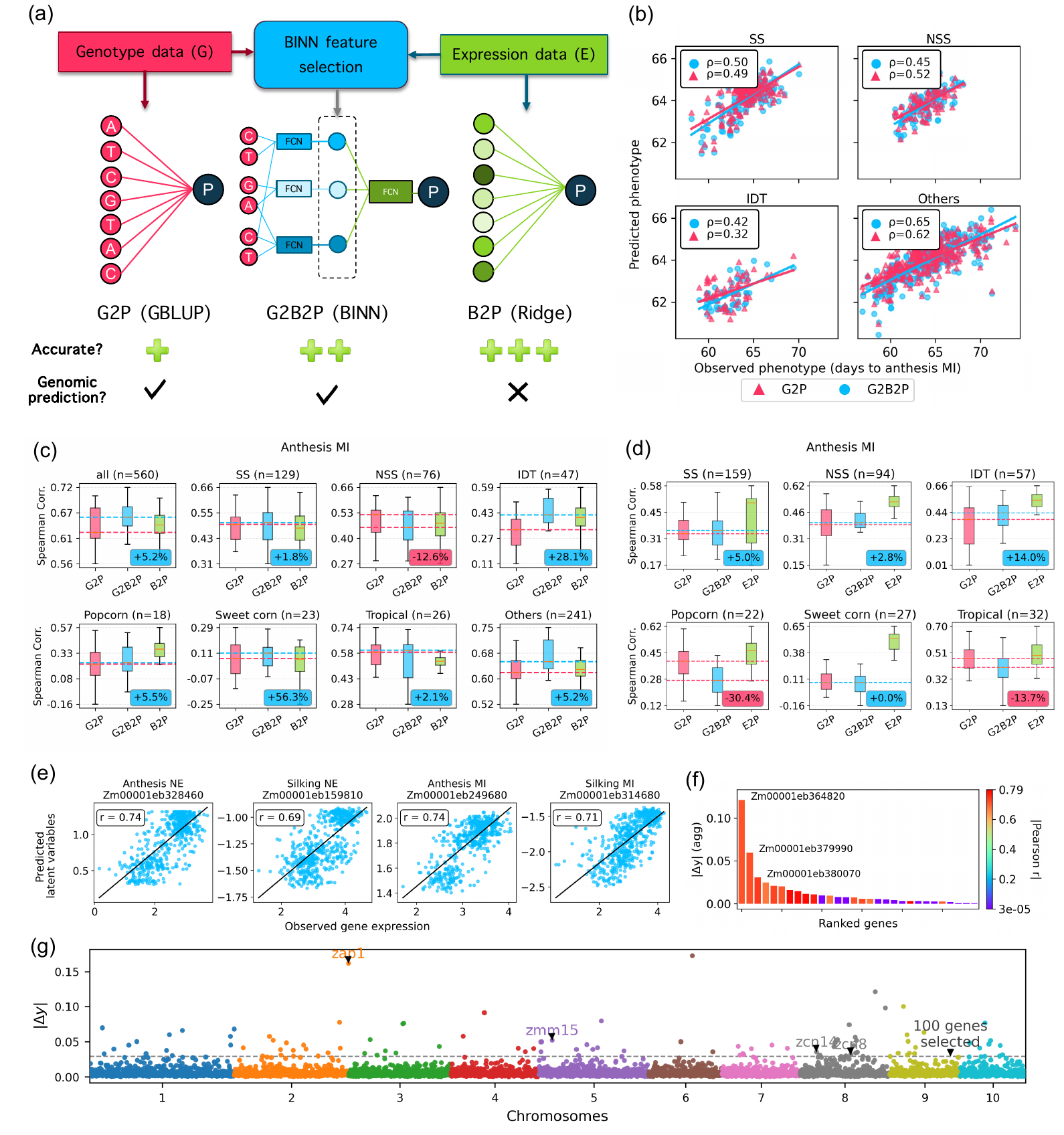}
  \caption{\textbf{BINNs improve prediction accuracy and interpretability through utiliziation of gene expression in genotype to phenotype modeling.} (a) Schematic of the transcriptomics-derived BINN architecture. The SNP marker and gene expression data are utilized for feature selection, which serves to sparsify the connections between the input and intermediate layers of the BINN architecture. The marker data for each gene is passed through pathway subnetworks at the intermediate layer and the outputs from this layer is passed through a non-linear integrator network to predict the phenotype values. The G2P and B2P models are both linear models. (b) Predicted vs. observed phenotype days to anthesis in MI across four subpopulations (SS, NSS, IDT, and Others). Points show G2P (GBLUP on genotype; pink triangles) and G2B2P (BINN with expression-informed sparsity; blue circles). An ordinary least-squares (OLS) fit is overlaid and the Spearman correlation ($\rho$) is reported in the legend. (c) Test Spearman correlation distributions for Silking NE, comparing three predictive models (G2P, G2B2P and B2P: Ridge regression on expression, green) across five independent 20/80\% train/test splits, each with five‑fold cross‑validation to capture both split‑level and CV‑level variability. Each subplot shows results for all lines pooled (“all”) and for each of the seven distinct subpopulations (SS, NSS, IDT, Popcorn, Sweet corn, Tropical and Others). The pink dashed line marks the median of the G2P baseline; the dashed line for G2B2P is colored green when its median exceeds that baseline or red when it falls below. Legends report the median percent change of G2B2P relative to G2P and titles show the number of test samples. (d) Test Spearman correlation distributions for leave-one-population-out experiments for Silking NE, comparing the same three predictive models. Each subplot shows results for the predictions on six distinct subpopulations using the models that are trained by leaving out the corresponding subpopulation from the training data. (e) Predicted latent variables vs.\ observed gene expression for four representative high-correlation genes per phenotype. (f) Aggregated absolute phenotypic change for 30 representative genes - 15 with high absolute Pearson correlation and 15 close to zero. (g) Aggregated absolute phenotypic change per perturbed gene, with a threshold capturing the BINN-derived 100 most significant genes which includes zap1, zmm15, zcn14 and zcn8.}
  \label{fig:schematic-twas}
\end{figure}

\textbf{Gene-expression-informed BINNs improve spearman rank correlations over G2P models across and within populations in maize field trials.} BINNs embed expression-derived sparsity into a genotype-to-biological-knowledge-to-phenotype ``G2B2P'', where biological-knowledge is gene expression data, network which delivers superior predictive performance compared to genotype-only GBLUP models, under sparse-data conditions (Figure \ref{fig:schematic-twas}b). We use the maize TWAS dataset from Torres-Rodríguez et al.~\cite{torres2024population}, which provides matched genotypes, RNA-seq expression profiles, and flowering-time phenotypes: days-to-anthesis and days-to-silking in Nebraska and Michigan (see Figure~\ref{fig:schematic-twas}b). In five independent 20\%/80\% train/test splits with five-fold cross-validation on each training set, BINN achieves higher Spearman correlations than the GBLUP baseline, both when pooling all 693 lines and within each of the seven heterotic subpopulations. BINNs outperformed GBLUP in the majority of pairwise comparisons, and the advantage was statistically significant on a paired t-test ($p=2.23\times10^{-6}$). Here, we report representative results for silking NE as shown in Figure~\ref{fig:schematic-twas}c; complete results for all phenotypes are provided in the Supplementary Figure 5. Notably, the most pronounced and consistent improvements occurred within individual subpopulations, the most critical and challenging scenario for GS, demonstrating BINN’s ability to capture subtle, group-specific genetic variation.

\textbf{BINNs leverage domain knowledge to pinpoint key genes and derive SNP–gene connections that both shrink network size while preserving nonlinear modeling power.} 
Models that embed domain knowledge have frequently demonstrated similar or superior predictive power compared to purely genotype-based approaches \cite{azodi2020transcriptome}. In the maize TWAS application, we observe that expression-based domain-to-phenotype (B2P) models outperform conventional G2P baselines across most evaluation settings and BINNs can translate this predictive strength into the G2P setting. Transcriptomic profiles inform the degree of architectural sparsity without ever feeding raw expression values into the prediction network at inference. Specifically, we first fit an Elastic Net model to each flowering‐time trait—tuning its L1/L2 penalty via cross‐validation—and selected the genes with nonzero coefficients, noting that the feature selection outcome was particularly sensitive to the specific cross-validation split. We then carried out eQTL mapping on these candidates to nominate their top SNP markers, which defined the sparse SNP→gene mask that shapes our G2B2P BINN architecture (Figure \ref{fig:schematic-twas}a). As shown in Table \ref{tab:binn_l1_sweep_all_traits}, a sweep of the Elastic Net’s L1 ratio revealed that retaining approximately 1,000 genes per trait and split offers the best trade-off between model compactness and accuracy, recovering the major regulators identified in the original study. An important limitation to this approach is that if B2P underperforms G2P, BINNs are unlikely to provide an advantage, since intermediate feature selection would not add predictive power.

\begin{table}[t]
\centering
\footnotesize
\caption{Test set Spearman rank correlation of BINNs across all populations and cross-validation splits for various Elastic Net L1 ratios used to select genes for the G2B2P mask for all four phenotypes. The best setting for each trait is bolded.}
\label{tab:binn_l1_sweep_all_traits}
\begin{tabular}{@{}l c c c c c@{}}
\toprule
\textbf{L1 ratio} & \textbf{\# genes} & \textbf{Anthesis NE} & \textbf{Anthesis MI} & \textbf{Silking NE} & \textbf{Silking MI} \\
\midrule
0.05  & 1541–2251  & 0.650 $\pm$ 0.034  & 0.652 $\pm$ 0.046  & 0.630 $\pm$ 0.041  & 0.649 $\pm$  0.036         \\
\textbf{0.10} & \textbf{848–1271} & \textbf{0.655 $\pm$ 0.037} & \textbf{0.663  $\pm$ 0.034} & \textbf{0.632 $\pm$ 0.040} & \textbf{0.660  $\pm$ 0.044} \\
0.20  & 467–696  & 0.605 $\pm$ 0.151    & 0.656 $\pm$ 0.038    & 0.623 $\pm$ 0.043   &  0.662 $\pm$ 0.038  \\
0.50  & 132-301  & 0.595 $\pm$ 0.143 & 0.633 $\pm$ 0.053  & 0.593 $\pm$ 0.127  & 0.640 $\pm$ 0.040              \\
1.00  & 50-131  & 0.597 $\pm$ 0.047  & 0.596 $\pm$ 0.047 &  0.547 $\pm$ 0.114 &  0.593 $\pm$ 0.059               \\
\bottomrule
\end{tabular}
\end{table}

\textbf{BINNs exploit shared expression–trait mechanisms beyond marker structure to improve generalization but rigid priors can limit flexibility in genetically distant populations.} It is well established that marker data are strongly tied to population structure whereas expression data are often tissue and environment-specific and less tightly tied to ancestry \cite{torres2025evolving}. Removing systematically distinct populations from the BINN training set revealed a clear pattern in model transferability. Expression data overall offers improved cross-population generalization compared to marker-based models as expression reflects functional output of many regulatory layers that can “normalize” some of the divergence in raw markers. However, BINN’s biologically constrained architecture can only capitalize on this advantage when the causal paths through biology are shared between train and test lines. Evidently, when one of the mainstream heterotic groups such as SS, NSS, or IDT was held out, BINN maintained robust generalization performance (see Figure \ref{fig:schematic-twas}d). This is particularly important for large-scale breeding programs, which rely heavily on these temperate pools for developing elite germplasm and driving genetic gain. 
In contrast,  when genetically more distant populations were excluded from training, such as sweet corn or tropical, which represent cases where regulatory programs diverge substantially from temperate pools \cite{wu2015analysis}, BINN appeared to overfit to pathway patterns dominated by the larger heterotic groups thereby limiting its ability to adapt. Thus, while BINNs excel when shared biological mechanisms exist, overly rigid priors limit its flexibility in zero-shot learning scenarios. However, this issue diminishes as domain knowledge improves: the more precise and mechanistically grounded it is, the larger the performance gains across tasks (see section \ref{sec:meta}).

\textbf{Sensitivity analysis reveals nonlinear genetic contributors that TWAS/GWAS may fail to capture.} A key advantage of BINNs lies in their interpretability, i.e., the ability of trained models to elucidate how specific biological entities influence intermediate traits and ultimately shape the phenotype. Traditional BINNs with single-edge weights offer direct interpretability, as each parameter corresponds to a distinct marker–gene or gene–trait connection. In our implementation, we enhance model expressivity by replacing these single-edge weights with fully connected layers, sacrificing one-to-one parameter interpretability but enabling richer representations of nonlinear dependencies. To assess whether BINNs can identify biologically meaningful genes, we developed and conducted a sensitivity analysis (presented in Algorithm \ref{alg:sensitivity}) across all pathway subnetworks. Specifically, we perturbed each pathway’s latent variable in both directions and quantified the resulting change in the predicted phenotype, ranking genes by their aggregated effect. Figure~\ref{fig:schematic-twas}e summarizes this analysis, showing that BINN recovers several TWAS-significant genes reported in Torres-Rodríguez et al.~\cite{torres2024population}, such as the previously characterized zcn8 as well as zap1, zmm15 and zcn14, but also highlights additional candidates that may participate in epistatic interactions shaping phenotypic response. These genes may be overlooked by traditional approaches like TWAS, which rely on marginal gene-trait associations, whereas BINN’s end-to-end training captures predictive, nonlinear, and combinatorial effects beyond the reach of standard linear models. Interestingly, we found that the genes most sensitive to phenotype perturbations also exhibit strong latent-expression correlations (Figure \ref{fig:schematic-twas}e,f). This suggests that BINNs preferentially learn biologically grounded representations, linear or nonlinear, aligned with known mechanisms.

\subsection{Metabolomics-derived BINN} \label{sec:meta}

We now demonstrate that BINNs offer greater potential in both accuracy and interpretability for genomic design when \textbf{established domain knowledge} is incorporated. This contrasts with the previous example, which relied solely on intermediate experimental data. To illustrate, we use a synthetic metabolomics problem in which hormone–sugar interactions are formalized through an ODE model, creating a clean G2P nonlinear test bed for evaluating BINNs. A motivating example comes from axillary bud outgrowth—the switch-like process underlying shoot branching—arising from interactions among auxin (A), cytokinin (CK), strigolactone (SL), and sucrose (S) in a minimal network \cite{bertheloot2020sugar}, later extended with genotype-to-metabolite variation at multiple causal loci \cite{powell2022investigations}. The domain knowledge we embed is simple but definitive: gene–metabolite, metabolite–metabolite, and metabolite–phenotype interactions. Importantly, the BINN is not provided with the full ODE details (the magnitudes of effects, equation structures, or nonlinearities) but only the associations between layers. To avoid an artificially “easy” setting and to test robustness, we perturb the simulated data by adding 5\% noise to intermediate traits and 10\% noise to phenotypes. This controlled setting enables us to rigorously test the BINN methodology: evaluating performance under both plentiful and sparse data conditions, examining its ability to recover known causal dynamics, and exploring extensions such as custom loss functions applied to partially observed intermediates.

\begin{figure}[ht]
  \centering
\includegraphics[width=0.8\linewidth]{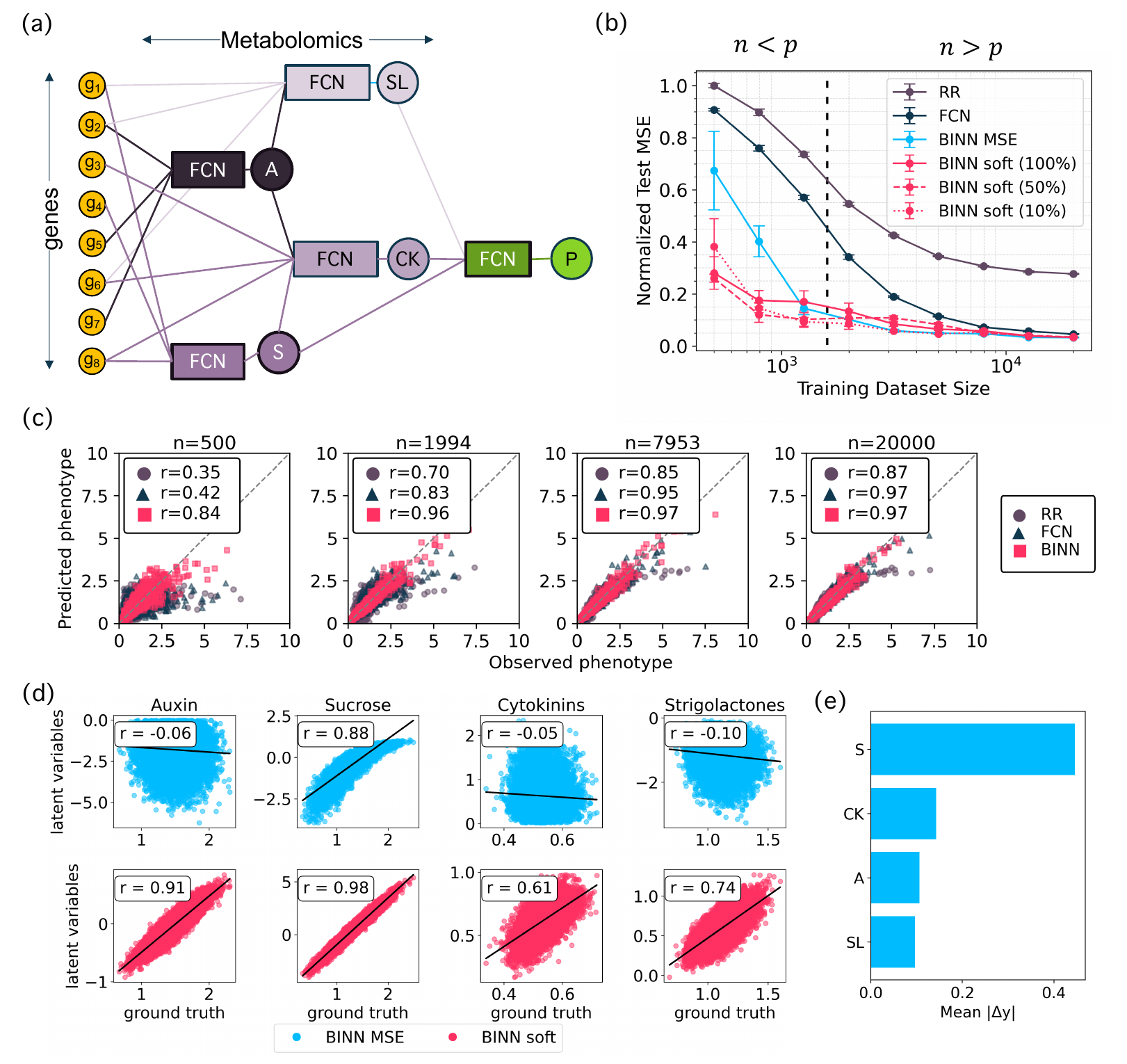}
  \caption{\textbf{BINNs significantly outperform baselines in the sparse data limit $n<p$}. (a) Schematic of the BINN architecture for the shoot‐branching network. Gene inputs are routed through four biologically‐annotated pathway subnetworks corresponding to auxin (A), sucrose (S), cytokinins (CK), and strigolactones (SL), then combined in a final integrator to predict bud‐outgrowth time. (b) Test‐set performance (MSE) for six models: ridge regression (RR), a generic fully‐connected network (FCN), BINN trained with standard MSE (BINN‐MSE), and BINN trained with the biologically‐informed soft‐constraint loss (BINN‐soft) with a varying fraction of known intermediate trait measurements (100\%, 50\% and 10\%), evaluated across nine training‐set sizes logarithmically spaced between 500 and 20,000 samples. Boxplots display the distribution across five random initializations. The vertical black dashed line at $n=1,600$ denotes the transition from sparse ($n<p$) to plentiful ($n>p$) data regimes. (c) Predicted vs observed phenotype across four training sizes ($n$ = 500, 1,994, 7,953, 20,000) for three models: RR (purple circles), FCN (black triangles), and BINN (red squares). Each panel shows the identity line (y = x), and in-panel legends report Pearson correlation (r) for each model. (d) Scatter plots of predicted versus true latent values for each intermediate trait (A, S, CK, SL) under the BINN MSE (blue) and the BINN soft model (red). The fitted regression line and Pearson correlation coefficient, r, are overlaid. (e) Aggregated absolute phenotypic change per perturbed intermediate trait.}
  \label{fig:Results-shoot-branching}
\end{figure}

\textbf{BINNs improve predictions under sparse‐data regimes.} In the genotype-to-metabolites-to-phenotype setup, BINN embeds ODE-derived metabolite pathways and constrains gene inputs through known gene–metabolite links. (Figure \ref{fig:Results-shoot-branching}a). Evaluated across nine geometrically-spaced training sizes from 500 to 20,000 lines, with five random splits, the standard MSE-trained BINN consistently outperformed Ridge regression and an unconstrained fully-connected network (FCN) in the sparse-data regime, i.e., when the number of training lines was smaller than the input dimensionality (Figure \ref{fig:Results-shoot-branching}b,c). As expected, when sample size increases, BINN performance approaches that of the FCN, indicating that the pathway-guided sparsity yields a better bias–variance trade-off when data are limited, while also preserving nonlinear expressivity as data grow. This demonstrates that BINNs can faithfully capture the intrinsic nonlinearity encoded in the underlying ODE dynamics across the entire data range, an aspect fundamentally beyond the reach of linear models like RR and GBLUP in the data-plentiful limit ($n\gg p$) and common neural networks in the data-sparse limit ($n\ll p$). 

\textbf{Biology-informed loss functions, applied to intermediate variables, improves prediction accuracy and recovery of pathway dynamics with sparse intermediate labels.} In many systems, intermediate traits can be experimentally measured, and this information can be used to better anchor latent variables to pathway dynamics. To this end, we introduce a soft-constraint loss function (i.e. Pearson loss) that explicitly encourages latent outputs to correlate with ground-truth intermediate trait values. In realistic scenarios, budget or experimental constraints often limit the amount of intermediate data that can be collected. To reflect this, we evaluate the model under three conditions: 100\%, 50\%, and 10\% of lines with known labels. This design preserves genotype-only inference while allowing for complete or partial pathway supervision during training. The soft-constraint approach (BINN soft), achieves the lowest test MSE in the small-data regime, and performs comparably to standard BINNs with larger datasets (Figure~\ref{fig:Results-shoot-branching}b). Notably, the performance of is largely insensitive to the proportion of intermediate data available, indicating that even sparse biological supervision enables the network to recover the underlying hormone–sugar dynamics. In other words, even a small amount of alignment is sufficient to boost downstream phenotypic predictions.

\textbf{BINN architectures provide a latent variable mechanism for interpretability, uncovering the critical importance of sucrose's impact on outgrowth.} When we analyze the biologically-informed latent variables via scatter‐plot diagnostics on 20,000 held-out test lines, distinct pathway-level patterns emerge. We ran this experiment under both the standard MSE objective (BINN MSE) and a custom soft-constraint MSE (BINN soft). As expected, the soft-constraint setup (see Figure \ref{fig:Results-shoot-branching}d (second row)) yields stronger correlations between latent variables and ground-truth metabolites, since the model was partially supervised to do so. However, an interesting observation comes from Figure \ref{fig:Results-shoot-branching}d (first row): with the standard BINN MSE, most hormone pathways show little correlation with the ground truth, yet sucrose displays a remarkably strong alignment. As shown in Figure \ref{fig:Results-shoot-branching}e, a post-hoc sensitivity analysis on the four intermediate traits demonstrated that the phenotype is more sensitivity to perturbations in sucrose, highlighting its importance. Indeed Bertheloot et al.~\cite{bertheloot2020sugar}, designed their experiments to show the critical importance of sucrose in this biological example. These results illustrate how BINNs can learn nonlinear relationships and surface biologically meaningful variables when appropriate inductive bias is added, achieving interpretability through both explicit alignment losses and emergent signals in the unconstrained setting.

\section{Discussion}

Biologically informed neural networks provide the flexibility to incorporate ever-growing omics datasets for genomic selection, embedding intricate biological mechanisms and delivering improved accuracy and generalization. By using intermediate molecular information to shape network sparsity and connectivity, BINNs convert domain knowledge into inductive bias that both stabilizes learning in sparse-data settings and yields pathway-level latent variables for interpretation. In this work we benchmark BINNs on two concrete tasks: an ODE-grounded, synthetic shoot-branching problem and a real maize TWAS dataset, to test whether they deliver \emph{accurate yet deployable} genomic predictions. Across both applications, BINN outperforms baseline Ridge/FCN/GBLUP models in the \emph{sparse-data regime}, remains stable under noise, and requires only genotypes at inference. Importantly, the trained model can be leveraged through post hoc sensitivity analysis to identify biologically relevant entities driving the phenotype, offering a nonlinear alternative to traditional GWAS and TWAS approaches.

Our BINN extension is built first and foremost for genomic-selection practicality. The primary focus of existing BINN implementations in the literature has not considered GS or plant breeding utility and while although multiple studies have demonstrated the strong predictive power of omics data, it is not always clear how to translate these insights into models that rely exclusively on genomic inputs for GS.  Our goal is to reframe the use of BINNs for GS that would make it relevant for many practical use cases in plant breeding. Full interpretability of existing models facilitates causal insight by directly linking predicted phenotypes to specific biological entities, but this rigid structure can constrain expressivity and thus limit predictive accuracy. A related challenge in plant breeding is scaling such models: breeders need decision‐support tools that deliver accurate predictions without incurring the time and cost of routine omics measurements. In this work, we present a new biology-aware design and demonstrate its ability to improve predictive performance and practicality for GS, offering a hopeful path toward more accurate and scalable crop improvement. This is achieved by the following  key features:





\textbf{BINNs do not need omics data at test time making them practical for the design phase where they are most needed and impactful.} As discussed, omics data carry high predictive power and, when leveraged appropriately, can enhance models that rely solely on genotype data. However, a critical constraint in this setting is that any omics data downstream of the genotype will not be available at deployment time, therefore such information must be leveraged only during training to shape model structure and guide learning, without being directly used at inference. Our BINN implementation folds omics data into inductive biases during training but never at inference allowing us to maintain full compatibility with standard GS workflows. Consistent with our out-of-distribution experiments, when the target population shares sufficient genetic background with the training cohort, these learned inductive biases transfer effectively, enabling reliable inference on new genotypes whereas gains naturally decrease for more divergent populations.


\textbf{Tunable sparsity of BINNs reduces overfitting risks, sensitivity to noise, and improved predictive performance.} We derive each omics‐layer mask by applying feature‐selection and association analyses to nominate candidate causal features across the dataset and match model architecture with trait complexity. Those candidates determine the sparse connectivity of the BINN model, allowing it to flexibly span oligogenic traits, where a few high-impact modules drive most of the signal, and highly polygenic traits, where predictive power emerges from many small‐effect contributors, simply by tuning the model sparsity. The tuned BINN models developed for two diverse applications delivered superior performance under sparse data, achieving up to a 56\% gain in test-set rank correlation and a 75\% reduction in predictive error relative to G2P baselines, within and across populations.

\textbf{Layered nonlinearity increases expressivity and allows BINNs to capture epistasis and nonlinear interactions.} BINNs demonstrated superior performance when learning the nonlinear metabolics problem, especially under sparse data. A key distinction of our approach lies in the balance between expressivity and interpretability. Unlike existing BINNs that assign each hidden neuron to a specific biological entity such as a gene, pathway, or protein, enabling direct read-off of effect weights, we employ flexible fully connected modules at each omics layer, enabling the network to learn rich, non-linear interactions among features while still honoring sparsity constraints derived from annotation even if individual hidden units no longer carry explicit biological labels. Each module’s small FCN can capture intra-pathway epistasis among genes (or other genetic regions, e.g. SNPs) and other local non-linear effects that single-weight connections miss. This is valuable when potential locus-locus interactions contribute to the phenotype. The final integrator network fuses these module outputs, capturing higher-order, between-gene interactions. 





\textbf{BINN latent variables offer opportunities to find hidden layer causal relationships given their stand in for biological domain knowledge.}
Even though we emphasize expressivity over built-in neuron-level interpretability, our BINN framework still offers opportunities for biological insight by uncovering causal relationships and enabling targeted sensitivity analyses that can inform biological understanding and decision-making. BINNs can prioritize intermediate traits by perturbing their corresponding latent variables and observing the resulting effect on the phenotype. In our metabolomics example, the phenotype showed the strongest sensitivity to perturbations of the sucrose latent variable, suggesting that the model captures pathway-relevant biological signals even under imperfect calibration. In the transcriptomics case, BINN not only recovers genes consistent with prior TWAS findings but also uncovers a range of additional candidates potentially influencing the phenotype through nonlinear, epistatic interactions. While these findings warrant further experimental validation, BINNs provide a promising framework for prioritizing gene candidates in downstream applications such as functional genomics and gene editing.

\textbf{BINNs are not without several limitations that must be considered when designing them.} 
Despite the promising performance of BINNs in both applications in this study there are several important limitations to consider. Introducing sparsity constraints as inductive biases can boost accuracy in data‑scarce settings, but the degree of sparsity must be carefully calibrated. Too little or too much undermines model behavior and thus, a systematic analysis to determine the optimal sparsity level is essential. Moreover, omics datasets are often noisy and heterogeneous, making it nontrivial to translate one or multiple –omics layers into an effective BINN architecture without risking mis‑specification. Like all ANN‑based approaches, BINNs demand  hyperparameter tuning and computational resources for proper calibration. 

One of the strengths of the BINN framework lies in its flexibility to incorporate different types of loss functions depending on the modeling objective, however, we did not find this improved results for the transcriptomics dataset. For example, one can augment the baseline predictive loss with biologically motivated constraints, such as pathway coherence penalties or sparsity-inducing terms, to encourage solutions that are both accurate and mechanistically plausible. We found that such auxiliary constraints can sometimes boost predictive accuracy, although the magnitude of improvement varies by application and requires careful balancing to avoid over-penalization. In our experiments, we also test a more rigid formulation by embedding a “hard” constrained MSE penalty designed to strongly enforce reconstruction consistency across latent pathways. However, this constraint did not yield improvements, suggesting that overly strict losses can reduce flexibility and hinder optimization. This highlights an important trade-off: while additional biological constraints can enhance performance and interpretability, they must be tuned with care.



To build on these findings and design BINNs in real‑world breeding programs, more studies most be taken to understand how BINNs perform across diverse applications and populations, with different domain knowledge, on varied phenotypes, and more. The ``sparse data'' regime varies by dataset size and population diversity, and the benefits of BINNs will depend on both trait complexity and crop species. While our work addressed relatively simple phenotypes (flowering time and time to bud outgrowth), systematic benchmarking across a broader spectrum of traits (e.g., from highly heritable, single‑gene characteristics to complex, polygenic attributes) will be essential to fully realize the potential of such models in crop improvement.

\begin{table}[ht]
  \centering
  \small
  \caption{Biological layers, their functions, and example data types.}
  \label{tab:bio-levels}
  \begin{tabular}{
      >{\raggedright\arraybackslash}p{3cm}
      >{\raggedright\arraybackslash}p{5cm}
      >{\raggedright\arraybackslash}p{4cm}
    }
    \toprule
    \textbf{Biological Level} & \textbf{Function} & \textbf{Example Data Type} \\
    \midrule
    \textbf{Genotype} & Genetic information underlying traits & SNP arrays, whole-genome sequencing \\
    Epigenetic Modifications & Influence gene regulation without changing DNA sequence & DNA methylation \\
    Gene Expression & Controls gene activity affecting traits & RNA-seq \\
    Protein Interaction Networks & Provide structural and signaling connections influencing outcomes & Proteomics \\
    Metabolites & Biochemical traits linking genes to observable traits & Metabolomics \\
    Regulatory Pathways & Control and coordinate gene expression & Pathway databases (e.g., KEGG) \\
    \textbf{Phenotype} & Observable traits resulting from genetic and environmental interactions & Yield, height, flowering time, etc. \\
    \bottomrule
  \end{tabular}
\end{table}



\textbf{Importantly, BINN gains depend on the quality of domain knowledge embedded in the model.} Purely data-driven setups, such as our transcriptomics-derived case, are constrained by the limits of experimental measurement: gene expression, while richer than genotype alone, is highly context-dependent (tissue, environment, sampling window) and thus noisy and heterogeneous, making performance sensitive to study design. By contrast, when mechanistic knowledge is sharper and more stable, illustrated by our metabolomics example, BINNs deliver markedly better accuracy in the sparse-data regime. This underscores a broader point: advancing fundamental biological understanding is not optional but enabling for practical GS models. Progress can come both from targeted experimental studies that elucidate pathways and from computational advances, e.g., genomic language models (gLMs)~\cite{benegas2023dna, benegas2025dna, nguyen2024hyenadna, nguyen2024sequence, benegas2025genomic} and functional genomics predictors~\cite{avsec2021effective,jaganathan2019predicting}, that infer function for genes lacking annotation, ultimately converting provisional signals into strong inductive biases we can embed in BINNs.

Future efforts should explore integrating environmental signals, e.g.\ weather time series, and further tailored strategies, such as integrating additional omics layers to more closely align BINN's causal pathways with mechanistic biology while maintaining a sound bias-variance trade-off. A practical challenge is that available omics are often heterogeneous, collected from different tissues, developmental stages, platforms, and cohort designs (single vs. multiple genotypes, narrow vs. diverse populations), making it nontrivial to decide how, or whether, to use them as priors. As summarized in Table~\ref{tab:bio-levels}, priors can be drawn from transcriptomics, proteomics, metabolite pathways, and curated interaction networks, enabling models that better reflect how genetic variation propagates through biological systems to affect traits. This can be facilitated by experimenting with alternative topologies (e.g., stacked-layer, or parallel-layer BINNs), and incorporating advanced neural modules to model pathway subnetworks more effectively. The goal should not be just higher accuracy, but robust, genotype-only predictors that surface pathway importance and translate into selection decisions. With these targeted extensions, BINNs are well-positioned to improve GS at scale.

\section{Methods}


Here we outline the BINN design used throughout before introducing the formalism. Our networks preserve nonlinear flexibility while enforcing biology-derived sparsity: genotype inputs are routed through pathway modules defined by masks built from prior knowledge (e.g., feature selection, association/eQTL hits, curated pathways, or ODE-based links). Each module is a small fully connected subnetwork that can model local nonadditivity and epistasis, and the module latents are fused by a final integrator to predict the phenotype. Intermediate omics measurements (e.g., expression, metabolites) inform the masks and, when available, can weakly supervise the latents via a soft constraint during training, but are never required at inference, preserving genotype-only deployment. The same template flexibly instantiates single-layer, stacked, staggered, or parallel multi-omics variants (see Supplementary Material) without changing the mathematical core. 

\subsection{Model Development with Biological Inductive Biases}
\label{sec:model}

We consider \(n\) samples with \(p\) SNP features collected in the matrix \(X\in\mathbb{R}^{n\times p}\) and a scalar phenotype vector \(y\in\mathbb{R}^n\).  Our BINN embeds \(L\) sequential omics layers between \(X\) and \(y\), each layer \(l\) producing a latent representation \(U^{(l)}\in\mathbb{R}^{n\times k_l}\) from its input \(U^{(l-1)}\).  We set \(U^{(0)} = X\) and denote the number of subnetworks, corresponding to biological entities such as genes, metabolites, or proteins, at layer \(l\) by \(k_l\).  

\paragraph{1. Pathway subnetworks.}  
At omics layer \(l\), prior biological knowledge (e.g.\ eQTL links, pathway membership) is encoded by a binary mask \(M^{(l)}\in\{0,1\}^{d_{l-1}\times k_l}\), where \(d_{l-1}=\text{dim}(U^{(l-1)})\).  Specifically,
\begin{equation}
M^{(l)}_{ij} =
\begin{cases}
1, & \text{if feature }i\text{ of }U^{(l-1)}\text{ maps to entity }j\text{ at layer }l,\\
0, & \text{otherwise}.
\end{cases}
\label{eq:mask_multilayer}
\end{equation}
For each of the \(k_l\) entities we define a subnetwork \(f^{(l)}_{j}\) that processes only the selected inputs \(U^{(l-1)}M^{(l)}_{:,j}\).  These subnetworks may have multiple hidden layers with sigmoid activations \(\sigma(u)=1/(1+e^{-u})\) to capture Hill‐type response typical in biological systems.  We then concatenate their outputs into
\begin{equation}
T^{(l)} = \bigl[f^{(l)}_{1}(U^{(l-1)}M^{(l)}_{:,1}), \dots, f^{(l)}_{k_l}(U^{(l-1)}M^{(l)}_{:,k_l})\bigr]\in\mathbb{R}^{n\times k_l},
\label{eq:pathway_multilayer}
\end{equation}
and set \(U^{(l)} = T^{(l)}\).

\paragraph{2. Residual network.}  
To capture genetic effects not explained by any pathway subnetworks, we apply an unconstrained residual network \(g_r\) (with parameters \(\theta_r\)) to the subset of SNPs that are not connected in any mask \(M^{(l)}\). Denoting this unannotated SNP matrix by \(X_{\mathrm{res}}\subset X\), the residual network directly predicts a scalar residual phenotype:
\begin{equation}
  r = g_{r}\bigl(X_{\mathrm{res}};\,\theta_{r}\bigr)
  \;\in\;\mathbb{R}^{n},
  \label{eq:residual_scalar}
\end{equation}
ensuring that \(r\) captures variation attributable to SNPs outside known biological pathways.

\paragraph{3. Final integrator network.}  
After the last omics layer produces \(U^{(L)}\in\mathbb{R}^{n\times k_L}\), we concatenate it with the residual output \(R\in\mathbb{R}^{n\times h_r}\):
\begin{equation}
Z = [\,U^{(L)},\,R\,]\;\in\;\mathbb{R}^{n\times (k_L + h_r)}.
\label{eq:integrator_input_multilayer}
\end{equation}
A final feed‐forward network \(h\) with parameters \(\theta_f\) then maps \(Z\) to the predicted phenotype
\begin{equation}
\hat y = h(Z;\theta_f)\;\in\;\mathbb{R}^n.
\label{eq:integrator_multilayer}
\end{equation}
which allows for the recovery of potential cross-pathway interactions. 

All network parameters \(\{W^{(l)},b^{(l)}\}_{l=1}^L\), \(\theta_r\), and \(\theta_f\) are learned jointly by minimizing a suitable loss function (see Section~\ref{sec:loss}).  The masks \(M^{(l)}\) enforce biological sparsity, reduce overfitting, and ensure each subnetwork aligns with known genotype–intermediate trait relationships.  

Our BINN framework is highly modular: depending on the number, type, and interdependencies of available omics layers, the network can assume different connectivity patterns, ranging from single‐layer designs to staggered, stacked, or parallel architectures. A brief description of these variants is provided in the Supplementary Material.

\subsection{Custom Loss Functions for Guided Training}
\label{sec:loss}
BINNs are flexible and can be trained with any suitable loss function, ranging from conventional objectives to biologically informed criteria.

\paragraph{Standard mean squared error (MSE).}  
The simplest choice is the mean squared error between the true phenotype \(y\) and the prediction \(\hat y\):
\begin{equation}
\mathcal{L}_{\mathrm{MSE}}(y,\hat y)
= \frac{1}{n}\sum_{i=1}^n \bigl(y_i - \hat y_i\bigr)^2.
\label{eq:mse_loss}
\end{equation}

\paragraph{Biologically informed soft‐constrained loss.}  
To encourage the model’s intermediate representations to align with measured omics traits, we add a soft constraint based on correlation.  Let \(T^{(l)} = [T^{(l)}_{1},\dots,T^{(l)}_{n}]\) be the predicted latent values at layer \(l\) and \(Z^{(l)} = [Z^{(l)}_{1},\dots,Z^{(l)}_{n}]\) the corresponding ground‐truth intermediate measurements.  Define the sample Pearson correlation
\begin{equation}
\rho^{(l)} 
= \frac{\sum_{i=1}^n \bigl(T^{(l)}_{i} - \bar T^{(l)}\bigr)\,\bigl(Z^{(l)}_{i} - \bar Z^{(l)}\bigr)}
       {\sqrt{\sum_{i=1}^n (T^{(l)}_{i} - \bar T^{(l)})^2}\,
        \sqrt{\sum_{i=1}^n (Z^{(l)}_{i} - \bar Z^{(l)})^2}},
\end{equation}
where \(\bar T^{(l)}\) and \(\bar Z^{(l)}\) are the layer‐wise means.  The biologically informed loss is then
\begin{equation}
\mathcal{L}_{\mathrm{bio}}
= \mathcal{L}_{\mathrm{MSE}}(y,\hat y)
+ \lambda
  \sum_{l=1}^L \bigl[\,1 - \rho^{(l)}\bigr],
\label{eq:bio_loss}
\end{equation}
where \(\lambda>0\) is a hyperparameter balancing phenotype accuracy against intermediate‐trait alignment.  One could instead enforce an exact match via an auxiliary hard-constrained MSE \(\|T^{(l)} - Z^{(l)}\|_2^2\), but the correlation‐based soft constraint is less restrictive and allows the model greater expressive power.

This biologically informed loss is optional: users may omit the correlation term (\(\lambda=0\)) to recover the standard MSE objective, or tune \(\lambda\) to any desired level of guidance.

\subsection{Sensitivity Analysis for Latent Perturbations}
\label{sec:sensitivity}

To quantify the contribution of intermediate biological entities to the phenotype, we perform a post-hoc \emph{sensitivity analysis} across trained BINN ensembles. This approach assesses how controlled perturbations of individual latent variables influence the predicted phenotype, providing an interpretable measure of importance that generalizes across omics layers, phenotypes, and model instances.

\paragraph{Overview.}
Given a trained BINN ensemble consisting of $S$ outer splits and $F$ inner folds, let $\{\mathcal{M}_{s,f}\}_{s=1..S,\,f=1..F}$ denote the set of trained models for a phenotype of interest. Each model $\mathcal{M}_{s,f}$ contains intermediate latent representations $U^{(l)}$ at layer $l$, corresponding to biological entities such as genes, metabolites, or proteins. Sensitivity analysis probes each latent by replacing it with controlled constants and observing the resulting change in the model's mean predicted phenotype $\hat{y}$.

\paragraph{Step 1: Baseline computation.}
For each trained model, we first perform a forward pass on a held-out test dataset to compute the baseline predicted phenotype $\hat{y}_0$ and the corresponding latent activations $U^{(l)}$. The mean $\mu_j$ and standard deviation $\sigma_j$ of each latent dimension $u^{(l)}_j$ are computed across all samples in the dataset.

\paragraph{Step 2: Latent clamping and phenotype response.}
To evaluate the sensitivity of entity $j$, we perform two controlled perturbations by \emph{clamping} its latent activation to constant values across all samples:
\begin{equation}
u^{(l)}_j(\delta) = \mu_j + \delta \,\sigma_j,
\quad \delta \in \{+a,\,-a\},
\label{eq:latent_clamp}
\end{equation}
while all other latents remain unaltered. The perturbed latent is then propagated through the trained model to produce new mean phenotype predictions $\hat{y}^{(+)}$ and $\hat{y}^{(-)}$. The per-entity sensitivity under model $\mathcal{M}_{s,f}$ is defined as the symmetric mean absolute deviation:
\begin{equation}
\Delta y^{(s,f)}_j = \tfrac{1}{2}\left(|\hat{y}^{(+)} - \hat{y}_0| + |\hat{y}^{(-)} - \hat{y}_0|\right),
\label{eq:symmetric_sensitivity}
\end{equation}
capturing the magnitude of the phenotype’s response to both upward and downward perturbations.

\paragraph{Step 3: Aggregation across models.}
Sensitivities are aggregated across all ensemble members that include entity $j$ to obtain a robust importance estimate:
\begin{equation}
\bar{\Delta y}_j
= \frac{1}{|\mathcal{M}(j)|}
  \sum_{(s,f)\in\mathcal{M}(j)} \Delta y^{(s,f)}_j,
\label{eq:aggregate_sensitivity}
\end{equation}
where $\mathcal{M}(j)$ is the subset of trained models that contain latent $u^{(l)}_j$. The resulting $\bar{\Delta y}_j$ represents the average change in predicted phenotype magnitude when the latent corresponding to entity $j$ is clamped to values spanning its natural variability.

\paragraph{Step 4: Entity ranking and downstream analysis.}
Entities are ranked by their aggregated sensitivity $\bar{\Delta y}_j$, yielding a model-based importance profile for the phenotype under study. This ranking can identify key intermediate traits, benchmark model interpretability against known associations, and guide downstream analyses such as candidate gene selection, pathway enrichment, or gene-editing prioritization.

\begin{algorithm}[H]
\caption{Sensitivity Analysis Across BINN Ensembles}
\label{alg:sensitivity}
\begin{algorithmic}[1]
\Require Trained models $\{\mathcal{M}_{s,f}\}$ for a phenotype; perturbation scale $a$
\For{each outer split $s=1..S$}
  \For{each inner fold $f=1..F$}
    \State Load trained model $\mathcal{M}_{s,f}$
    \State Compute baseline mean phenotype $\hat{y}_0$ and latent statistics $(\mu_j,\sigma_j)$
    \For{each latent entity $j$}
      \State Clamp latent $u_j \leftarrow \mu_j \pm a\cdot\sigma_j$ (constant across all samples)
      \State Forward model to compute mean predictions $\hat{y}^{(+)}$ and $\hat{y}^{(-)}$
      \State Compute $\Delta y^{(s,f)}_j = \tfrac{1}{2}\left(|\hat{y}^{(+)} - \hat{y}_0| + |\hat{y}^{(-)} - \hat{y}_0|\right)$
    \EndFor
  \EndFor
\EndFor
\State Aggregate $\bar{\Delta y}_j = \text{mean}_{(s,f)}(\Delta y^{(s,f)}_j)$ over all models containing $j$
\State Rank entities by $\bar{\Delta y}_j$
\end{algorithmic}
\end{algorithm}



\section{Data Availability}
All data generated for the shoot‑branching problem were produced in‑house by numerically integrating the ODE frameworks described in Powell et al.~\cite{powell2022investigations} and Betherloot et al.~\cite{bertheloot2020sugar}. For the maize TWAS analyses, we used the publicly released genotype–expression–phenotype dataset from Torres-Rodríguez et al.~\cite{torres2024population}, which is accessible via the original publication’s data archive. 

\section{Code Availability}
All simulation code, analysis scripts, ElasticNet‑derived gene lists, trained models, and code to generate the figures in this paper are available upon request and are provided for non-commercial research use only.

\section*{Acknowledgements} 
We would like to thank Megan Gillespie for supporting and championing this work. Nathan Springer for his countless hours in helping us link and communicate our ideas towards biological and genomic applications. Alexis Charalampopoulos for early ideation on the concept of BINNs. As well as several team members who have contributed substantial feedback throughout this development of this work: Jaclyn Noshay, Sarah Turner-Hissong, Fabiana Freitas Moreira, Zhangyue Shi, Rocio Dominguez Vidana, and Koushik Nagasubramanian.

\bibliographystyle{plain}  
\bibliography{main-ref} 

\begin{thebibliography}{10}

\bibitem{alemu2024genomic}
Admas Alemu, Johanna {\AA}strand, Osval~A Montesinos-Lopez, Julio~Isidro y~Sanchez, Javier Fernandez-Gonzalez, Wuletaw Tadesse, Ramesh~R Vetukuri, Anders~S Carlsson, Alf Ceplitis, Jose Crossa, et~al.
\newblock Genomic selection in plant breeding: Key factors shaping two decades of progress.
\newblock {\em Molecular Plant}, 17(4):552--578, 2024.

\bibitem{atwell2010genome}
Susanna Atwell, Yu~S Huang, Bjarni~J Vilhj{\'a}lmsson, Glenda Willems, Matthew Horton, Yan Li, Dazhe Meng, Alexander Platt, Aaron~M Tarone, Tina~T Hu, et~al.
\newblock Genome-wide association study of 107 phenotypes in arabidopsis thaliana inbred lines.
\newblock {\em Nature}, 465(7298):627--631, 2010.

\bibitem{avsec2021effective}
{\v{Z}}iga Avsec, Vikram Agarwal, Daniel Visentin, Joseph~R Ledsam, Agnieszka Grabska-Barwinska, Kyle~R Taylor, Yannis Assael, John Jumper, Pushmeet Kohli, and David~R Kelley.
\newblock Effective gene expression prediction from sequence by integrating long-range interactions.
\newblock {\em Nature methods}, 18(10):1196--1203, 2021.

\bibitem{azodi2019benchmarking}
Christina~B Azodi, Emily Bolger, Andrew McCarren, Mark Roantree, Gustavo de~Los~Campos, and Shin-Han Shiu.
\newblock Benchmarking parametric and machine learning models for genomic prediction of complex traits.
\newblock {\em G3: Genes, Genomes, Genetics}, 9(11):3691--3702, 2019.

\bibitem{azodi2020transcriptome}
Christina~B Azodi, Jeremy Pardo, Robert VanBuren, Gustavo de~Los~Campos, and Shin-Han Shiu.
\newblock Transcriptome-based prediction of complex traits in maize.
\newblock {\em The Plant Cell}, 32(1):139--151, 2020.

\bibitem{benegas2025dna}
Gonzalo Benegas, Carlos Albors, Alan~J Aw, Chengzhong Ye, and Yun~S Song.
\newblock A dna language model based on multispecies alignment predicts the effects of genome-wide variants.
\newblock {\em Nature Biotechnology}, pages 1--6, 2025.

\bibitem{benegas2023dna}
Gonzalo Benegas, Sanjit~Singh Batra, and Yun~S Song.
\newblock Dna language models are powerful predictors of genome-wide variant effects.
\newblock {\em Proceedings of the National Academy of Sciences}, 120(44):e2311219120, 2023.

\bibitem{benegas2025genomic}
Gonzalo Benegas, Chengzhong Ye, Carlos Albors, Jianan~Canal Li, and Yun~S Song.
\newblock Genomic language models: opportunities and challenges.
\newblock {\em Trends in Genetics}, 2025.

\bibitem{bertheloot2020sugar}
Jessica Bertheloot, Fran{\c{c}}ois Barbier, Fr{\'e}d{\'e}ric Boudon, Maria~Dolores Perez-Garcia, Thomas P{\'e}ron, Sylvie Citerne, Elizabeth Dun, Christine Beveridge, Christophe Godin, and Soulaiman Sakr.
\newblock Sugar availability suppresses the auxin-induced strigolactone pathway to promote bud outgrowth.
\newblock {\em New Phytologist}, 225(2):866--879, 2020.

\bibitem{breiman2001random}
Leo Breiman.
\newblock Random forests.
\newblock {\em Machine learning}, 45:5--32, 2001.

\bibitem{christensen2021genetic}
Ole~F Christensen, Vinzent B{\"o}rner, Luis Varona, and Andres Legarra.
\newblock Genetic evaluation including intermediate omics features.
\newblock {\em Genetics}, 219(2):iyab130, 2021.

\bibitem{clark2013genomic}
Samuel~A Clark and Julius van~der Werf.
\newblock Genomic best linear unbiased prediction (gblup) for the estimation of genomic breeding values.
\newblock {\em Genome-wide association studies and genomic prediction}, pages 321--330, 2013.

\bibitem{crossa2017genomic}
Jose Crossa, Paulino P{\'e}rez-Rodr{\'\i}guez, Javier Cuevas, Osval~A Montesinos-L{\'o}pez, Diego Jarqu{\'\i}n, Gustavo de~Los~Campos, Juan Burgue{\~n}o, Jos{\'e}~Manuel Gonz{\'a}lez-Camacho, Salvador P{\'e}rez-Elizalde, Yoseph Beyene, and Susanne Dreisigacker.
\newblock Genomic selection in plant breeding: methods, models, and perspectives.
\newblock {\em Trends in Plant Science}, 22(11):961--975, 2017.

\bibitem{elmarakeby2021biologically}
Haitham~A Elmarakeby, Justin Hwang, Rand Arafeh, Jett Crowdis, Sydney Gang, David Liu, Saud~H AlDubayan, Keyan Salari, Steven Kregel, Camden Richter, et~al.
\newblock Biologically informed deep neural network for prostate cancer discovery.
\newblock {\em Nature}, 598(7880):348--352, 2021.

\bibitem{endelman2011ridge}
Jeffrey~B Endelman.
\newblock Ridge regression and other kernels for genomic selection with r package rrblup.
\newblock {\em The plant genome}, 4(3), 2011.

\bibitem{gamazon2015gene}
Eric~R Gamazon, Heather~E Wheeler, Kaanan~P Shah, Sahar~V Mozaffari, Keston Aquino-Michaels, Robert~J Carroll, Anne~E Eyler, Joshua~C Denny, GTEx Consortium, Dan~L Nicolae, et~al.
\newblock A gene-based association method for mapping traits using reference transcriptome data.
\newblock {\em Nature genetics}, 47(9):1091--1098, 2015.

\bibitem{gao2023soydngp}
Pengfei Gao, Haonan Zhao, Zheng Luo, Yifan Lin, Wanjie Feng, Yaling Li, Fanjiang Kong, Xia Li, Chao Fang, and Xutong Wang.
\newblock Soydngp: a web-accessible deep learning framework for genomic prediction in soybean breeding.
\newblock {\em Briefings in bioinformatics}, 24(6):bbad349, 2023.

\bibitem{gianola2009additive}
Daniel Gianola, Gustavo de~Los~Campos, William~G Hill, Eduardo Manfredi, and Rohan Fernando.
\newblock Additive genetic variability and the bayesian alphabet.
\newblock {\em Genetics}, 183(1):347--363, 2009.

\bibitem{hartman2023interpreting}
Erik Hartman, Aaron~M Scott, Christofer Karlsson, Tirthankar Mohanty, Suvi~T Vaara, Adam Linder, Lars Malmstr{\"o}m, and Johan Malmstr{\"o}m.
\newblock Interpreting biologically informed neural networks for enhanced proteomic biomarker discovery and pathway analysis.
\newblock {\em Nature Communications}, 14(1):5359, 2023.

\bibitem{heffner2009genomic}
Elliot~L Heffner, Mark~E Sorrells, and Jean-Luc Jannink.
\newblock Genomic selection for crop improvement.
\newblock {\em Crop Science}, 49(1):1--12, 2009.

\bibitem{jacquin2016unified}
Laval Jacquin, Tuong-Vi Cao, and Nourollah Ahmadi.
\newblock A unified and comprehensible view of parametric and kernel methods for genomic prediction with application to rice.
\newblock {\em Frontiers in genetics}, 7:145, 2016.

\bibitem{jaganathan2019predicting}
Kishore Jaganathan, Sofia~Kyriazopoulou Panagiotopoulou, Jeremy~F McRae, Siavash~Fazel Darbandi, David Knowles, Yang~I Li, Jack~A Kosmicki, Juan Arbelaez, Wenwu Cui, Grace~B Schwartz, et~al.
\newblock Predicting splicing from primary sequence with deep learning.
\newblock {\em Cell}, 176(3):535--548, 2019.

\bibitem{korte2013advantages}
Arthur Korte and Ashley Farlow.
\newblock The advantages and limitations of trait analysis with gwas: a review.
\newblock {\em Plant methods}, 9:1--9, 2013.

\bibitem{ma2017deepgs}
Wenlong Ma, Zhixu Qiu, Jie Song, Qian Cheng, and Chuang Ma.
\newblock Deepgs: Predicting phenotypes from genotypes using deep learning.
\newblock {\em BioRxiv}, page 241414, 2017.

\bibitem{ma2022omicsgcn}
Xiaojie Ma et~al.
\newblock Omicsgcn: a multi-view graph convolutional network for multi-omics data integration.
\newblock {\em Bioinformatics}, 2022.

\bibitem{mcdonald2009ridge}
Gary~C McDonald.
\newblock Ridge regression.
\newblock {\em Wiley Interdisciplinary Reviews: Computational Statistics}, 1(1):93--100, 2009.

\bibitem{meuwissen2001prediction}
Theo~HE Meuwissen, Ben~J Hayes, and ME1461589 Goddard.
\newblock Prediction of total genetic value using genome-wide dense marker maps.
\newblock {\em genetics}, 157(4):1819--1829, 2001.

\bibitem{montesinos2022bayesian}
Osval~Antonio Montesinos~L{\'o}pez, Abelardo Montesinos~L{\'o}pez, and Jose Crossa.
\newblock Bayesian genomic linear regression.
\newblock In {\em Multivariate Statistical Machine Learning Methods for Genomic Prediction}, pages 171--208. Springer, 2022.

\bibitem{montesinos2021review}
Osval~Antonio Montesinos-L{\'o}pez, Abelardo Montesinos-L{\'o}pez, Paulino P{\'e}rez-Rodr{\'\i}guez, Jos{\'e}~Alberto Barr{\'o}n-L{\'o}pez, Johannes~WR Martini, Silvia~Berenice Fajardo-Flores, Laura~S Gaytan-Lugo, Pedro~C Santana-Mancilla, and Jos{\'e} Crossa.
\newblock A review of deep learning applications for genomic selection.
\newblock {\em BMC genomics}, 22(1):19, 2021.

\bibitem{nguyen2024sequence}
Eric Nguyen, Michael Poli, Matthew~G Durrant, Brian Kang, Dhruva Katrekar, David~B Li, Liam~J Bartie, Armin~W Thomas, Samuel~H King, Garyk Brixi, et~al.
\newblock Sequence modeling and design from molecular to genome scale with evo.
\newblock {\em Science}, 386(6723):eado9336, 2024.

\bibitem{nguyen2024hyenadna}
Eric Nguyen, Michael Poli, Marjan Faizi, Armin Thomas, Michael Wornow, Callum Birch-Sykes, Stefano Massaroli, Aman Patel, Clayton Rabideau, Yoshua Bengio, et~al.
\newblock Hyenadna: Long-range genomic sequence modeling at single nucleotide resolution.
\newblock {\em Advances in neural information processing systems}, 36, 2024.

\bibitem{nguyen2020deepprog}
Tran Nguyen et~al.
\newblock Deepprog: an ensemble of deep-learning and machine-learning models for prognosis prediction using multi-omics data.
\newblock {\em Nature Communications}, 2020.

\bibitem{ogutu2012genomic}
Joseph~O Ogutu, Torben Schulz-Streeck, and Hans-Peter Piepho.
\newblock Genomic selection using regularized linear regression models: ridge regression, lasso, elastic net and their extensions.
\newblock In {\em BMC proceedings}, volume~6, pages 1--6. Springer, 2012.

\bibitem{powell2022investigations}
Owen~M Powell, Francois Barbier, Kai~P Voss-Fels, Christine Beveridge, and Mark Cooper.
\newblock Investigations into the emergent properties of gene-to-phenotype networks across cycles of selection: a case study of shoot branching in plants.
\newblock {\em in silico Plants}, 4(1):diac006, 2022.

\bibitem{raissi2019physics}
Maziar Raissi, Paris Perdikaris, and George~E Karniadakis.
\newblock Physics-informed neural networks: A deep learning framework for solving forward and inverse problems involving nonlinear partial differential equations.
\newblock {\em Journal of Computational physics}, 378:686--707, 2019.

\bibitem{riedelsheimer2012genomic}
Christian Riedelsheimer, Angelika Czedik-Eysenberg, Christoph Grieder, Jan Lisec, Frank Technow, Ronan Sulpice, Thomas Altmann, Mark Stitt, Lothar Willmitzer, and Albrecht~E Melchinger.
\newblock Genomic and metabolic prediction of complex heterotic traits in hybrid maize.
\newblock {\em Nature genetics}, 44(2):217--220, 2012.

\bibitem{ritchie2015methods}
Marylyn~D Ritchie, Emily~R Holzinger, Ruowang Li, Sarah~A Pendergrass, and Dokyoon Kim.
\newblock Methods of integrating data to uncover genotype--phenotype interactions.
\newblock {\em Nature Reviews Genetics}, 16(2):85--97, 2015.

\bibitem{torres2025evolving}
J~Vladimir Torres-Rodr{\'\i}guez, Delin Li, and James~C Schnable.
\newblock Evolving best practices for transcriptome-wide association studies accelerate discovery of gene-phenotype links.
\newblock {\em Current Opinion in Plant Biology}, 83:102670, 2025.

\bibitem{torres2024population}
J~Vladimir Torres-Rodr{\'\i}guez, Delin Li, Jonathan Turkus, Linsey Newton, Jensina Davis, Lina Lopez-Corona, Waqar Ali, Guangchao Sun, Ravi~V Mural, Marcin~W Grzybowski, et~al.
\newblock Population-level gene expression can repeatedly link genes to functions in maize.
\newblock {\em The Plant Journal}, 119(2):844--860, 2024.

\bibitem{van2021gennet}
Arno van Hilten, Steven~A Kushner, Manfred Kayser, M~Arfan Ikram, Hieab~HH Adams, Caroline~CW Klaver, Wiro~J Niessen, and Gennady~V Roshchupkin.
\newblock Gennet framework: interpretable deep learning for predicting phenotypes from genetic data.
\newblock {\em Communications biology}, 4(1):1094, 2021.

\bibitem{van2024phenotype}
Arno van Hilten, Jeroen van Rooij, M~Arfan Ikram, Wiro~J Niessen, Joyce~BJ van Meurs, and Gennady~V Roshchupkin.
\newblock Phenotype prediction using biologically interpretable neural networks on multi-cohort multi-omics data.
\newblock {\em NPJ systems biology and applications}, 10(1):81, 2024.

\bibitem{varshney2021genomic}
Rajeev~K Varshney, Manish Roorkiwal, and Mark~E Sorrells.
\newblock Genomic-enabled prediction models for improving crop productivity.
\newblock {\em Trends in Plant Science}, 26(6):575--587, 2021.

\bibitem{wainberg2019opportunities}
Michael Wainberg, Nasa Sinnott-Armstrong, Nicholas Mancuso, Alvaro~N Barbeira, David~A Knowles, David Golan, Raili Ermel, Arno Ruusalepp, Thomas Quertermous, Ke~Hao, et~al.
\newblock Opportunities and challenges for transcriptome-wide association studies.
\newblock {\em Nature genetics}, 51(4):592--599, 2019.

\bibitem{wang2021mogonet}
Dong et~al. Wang.
\newblock Mogonet integrates multi-omics data through graph convolutional networks.
\newblock {\em Nature Machine Intelligence}, 2021.

\bibitem{wang2025cropformer}
Hao Wang, Shen Yan, Wenxi Wang, Yongming Chen, Jingpeng Hong, Qiang He, Xianmin Diao, Yunan Lin, Yanqing Chen, Yongsheng Cao, et~al.
\newblock Cropformer: An interpretable deep learning framework for crop genomic prediction.
\newblock {\em Plant Communications}, 6(3), 2025.

\bibitem{wang2023dnngp}
Kelin Wang, Muhammad~Ali Abid, Awais Rasheed, Jose Crossa, Sarah Hearne, and Huihui Li.
\newblock Dnngp, a deep neural network-based method for genomic prediction using multi-omics data in plants.
\newblock {\em Molecular Plant}, 16(1):279--293, 2023.

\bibitem{washburn2025global}
Jacob~D Washburn, Jos{\'e}~Ignacio Varela, Alencar Xavier, Qiuyue Chen, David Ertl, Joseph~L Gage, James~B Holland, Dayane~Cristina Lima, Maria~Cinta Romay, Marco Lopez-Cruz, et~al.
\newblock Global genotype by environment prediction competition reveals that diverse modeling strategies can deliver satisfactory maize yield estimates.
\newblock {\em Genetics}, 229(2):iyae195, 2025.

\bibitem{wu2015analysis}
Xun Wu, Yongxiang Li, Xin Li, Chunhui Li, Yunsu Shi, Yanchun Song, Zuping Zheng, Yu~Li, and Tianyu Wang.
\newblock Analysis of genetic differentiation and genomic variation to reveal potential regions of importance during maize improvement.
\newblock {\em BMC plant biology}, 15(1):256, 2015.

\bibitem{yang2021mvae}
Xinyue Yang et~al.
\newblock mmvae: a multi-modal variational autoencoder framework for integrative analysis of multi-omics data.
\newblock {\em Bioinformatics}, 37(15):2151--2158, 2021.

\bibitem{yegnanarayana2009artificial}
Bayya Yegnanarayana.
\newblock {\em Artificial neural networks}.
\newblock PHI Learning Pvt. Ltd., 2009.

\end{thebibliography}



\section*{Supplementary material}

# Supplementary Materials for:

# *Biology-informed neural networks learn nonlinear representations from omics data to improve genomic prediction and interpretability*

Katiana Kontolati, Rini Jasmine Gladstone, Ian Davis, Ethan Pickering

## Overview

This document provides supplementary information supporting the findings and methodology of the main manuscript. It includes additional figures, tables, implementation details, and reproducibility notes for both the transcriptomics and metabolomics applications.

## 1 Metabolomics-derived BINN

Axillary bud outgrowth is the key developmental process driving shoot branching in plants. Its regulation arises from an intricate network of hormonal and sugar signals. In the framework of Bertheloot et al. [1], auxin (A) concentration produced at the shoot apex is transported basipetally and inhibits cytokinin (CK) synthesis while promoting strigolactone (SL) production. Cytokinin levels and strigolactone levels then act antagonistically to activate or repress bud growth, respectively, and sucrose availability (S) acts as an enabling cue. Bertheloot et al. showed that this minimal network captures the switch-like behavior of bud outgrowth across species and can be approximated as a system of ordinary differential equations (ODEs) [1]). Powell et al. [8] then extended this ODE framework by parameterizing each interaction term and intermediate-trait (hormones and sucrose) level with additive genetic values calculated based on the additive genetic effects at multiple causal loci.

The introduction of genetic variation into the shoot-branching network yields a fully specified G2P model, enabling in silico selection experiments that reveal how interactions among intermediate traits drive selection response, precisely the process breeders seek to optimize. Because this framework integrates genotype, metabolite levels, and phenotype, it functions as a biologically-informed model, embedding known genetic loci-metabolite relationships as inductive biases. Moreover, having access to the exact ODEs used to generate synthetic data gives us a ground-truth baseline against which to develop and rigorously benchmark our BINN architectures. Although the mapping from genotype to metabolite in this system is exactly specified, in real breeding populations we typically know only statistical associations and never the true functional forms. We therefore hypothesize that a BINN can recover

these hidden relationships, learning the underlying network dynamics directly from data and providing both accurate predictions and mechanistic insight.

## 1.1 Adapted system of equations

We generate a large synthetic dataset following the ODE-based shoot-branching network of Bertheloot et al. [1] extended by Powell et al. [8]. For reproducibility we outline the exact set of equations below. Given $N$ lines and $L$ causal loci per pathway, the procedure is:

**1. Simulate additive genetic effects.** Draw locus effects $\{\beta_j\}_{j=1}^{4L}$ from a standard normal distribution. Denote the resulting effect vector by

$$\boldsymbol{\beta} = (\beta_1, \ldots, \beta_{4L})^\top.$$

**2. Generate genotypes.** Form the genotype matrix

$$G \in \{0,1\}^{N \times 4L}, \quad G_{i,j} \sim \text{Bernoulli}(0.5),$$

where row $i$ represents line $i$ and columns $1 \ldots L,\ L+1 \ldots 2L,\ 2L+1 \ldots 3L,\ 3L+1 \ldots 4L$ correspond to auxin (A), sugar (S), cytokinin (CK), and strigolactone (SL) pathways.

**3. Compute additive pathway values.** For each pathway $p \in \{A, S, CK, SL\}$, let $\boldsymbol{\beta}^{(p)}$ and $G^{(p)}$ denote the corresponding length-$L$ subvector and submatrix. The raw genetic value for pathway $p$ in line $i$ is

$$g_i^{(p)} = \sum_{j=1}^{L} G_{i,j}^{(p)} \beta_j^{(p)}. \tag{1}$$

**4. Rescale to steady-state trait levels.** To match the ranges used by Powell et al. [8], each $g^{(p)}$ is linearly rescaled to an intermediate trait $x^{(p)}$:

$$x_i^{(p)} = \left( \frac{g_i^{(p)}}{2 \max_i |g_i^{(p)}|} + \tfrac{1}{2} \right) s_p + b_p, \tag{2}$$

with pathway-specific scale $s_p$ and offset $b_p$:

$$(s_A, b_A) = (2.5, 0), \quad (s_S, b_S) = (2.4, 0.1), \quad (s_{CK}, b_{CK}) = (0.6, 0.4), \quad (s_{SL}, b_{SL}) = (0.9, 0.2).$$

**5. Compute interaction terms and intermediate traits.** After computing raw genetic values $g^{(p)}$ via Eq. (1), we first linearly map auxin and sugar to obtain

$$x_i^{(A)} = g_i^{(A)}, \qquad\qquad x_i^{(S)} = g_i^{(S)}. \tag{3}$$

Cytokinin and strigolactone receive additional interaction adjustments as:

$$x_i^{(\text{CK})} = \frac{g_i^{(\text{CK})} \gamma_{\text{CK}\leftarrow\text{A}}^{(i)} + \gamma_{\text{CK}\leftarrow\text{S}}^{(i)}}{0.99}, \tag{4}$$

$$x_i^{(\text{SL})} = \frac{g_i^{(\text{SL})} + \gamma_{\text{SL}\leftarrow\text{A}}^{(i)}}{0.86}. \tag{5}$$

where $\gamma_{\text{CK}\leftarrow\text{A}}^{(i)}$, $\gamma_{\text{CK}\leftarrow\text{S}}^{(i)}$, and $\gamma_{\text{SL}\leftarrow\text{A}}^{(i)}$ are computed as

$$\gamma_{\text{CK}\leftarrow\text{A}}^{(i)} = \frac{1}{1 + 0.96\, x_i^{(\text{A})}}, \qquad \gamma_{\text{CK}\leftarrow\text{S}}^{(i)} = \frac{0.25\, (x_i^{(\text{S})})^2}{0.19 + (x_i^{(\text{S})})^2}, \tag{6}$$

$$\gamma_{\text{SL}\leftarrow\text{A}}^{(i)} = \frac{24.89\, (x_i^{(\text{A})})^2}{294.58 + (x_i^{(\text{A})})^2}. \tag{7}$$

**6. Integrator and phenotype.** The intermediate traits are passed throug the signal integrator term as

$$\gamma_{I\leftarrow\text{CK}}^{(i)} = \frac{1}{1 + 1000\, x_i^{(\text{CK})}}, \tag{8}$$

$$\gamma_{I\leftarrow\text{SL,S}}^{(i)} = 5.64\, \frac{(x_i^{(\text{SL})})^2}{1 + \left(0.00418 + 7.10\, (x_i^{(\text{S})})^2\right) (x_i^{(\text{SL})})^2}. \tag{9}$$

The composite integrator signal for line $i$ is

$$I_i = 0.33\; +\; \gamma_{I\leftarrow\text{CK}}^{(i)}\; +\; \gamma_{I\leftarrow\text{SL,S}}^{(i)}. \tag{10}$$

Time to bud outgrowth is then obtained by a linear transformation and clipping to remove negative days:

$$y_{\text{BO},i} = \min\{-3.5\, \min_j I_j + 3.5\, I_i,\; 8.3\}, \tag{11}$$

and lines with $y_{\text{BO},i} \leq 8.3$ are scored as having initiated bud outgrowth.

**7. Final dataset.** We generate the genotype matrix $X = G$, the phenotype vector $y_{\text{BO}}$, and the true intermediate traits $\{x^{(\text{A})}, x^{(\text{S})}, x^{(\text{CK})}, x^{(\text{SL})}\}$ for downstream model training and evaluation.

## 1.2 Data generation

We generated a synthetic dataset using the equations outlined in Section 1.1 using our own Python implementation. We consider $L = 400$ causal loci per pathway which results in $1{,}600$ loci in total. A large dataset of $N = 100{,}000$ lines was generated from which 10% was used for validation and 20% for testing fixed across models. The remaining 70% was used to randomly sample training subsets used in the experiments described below. Below, we show the distribution of the phenotype $y_{\text{BO}}$ and the four intermediate traits (A,S,CK,SL).

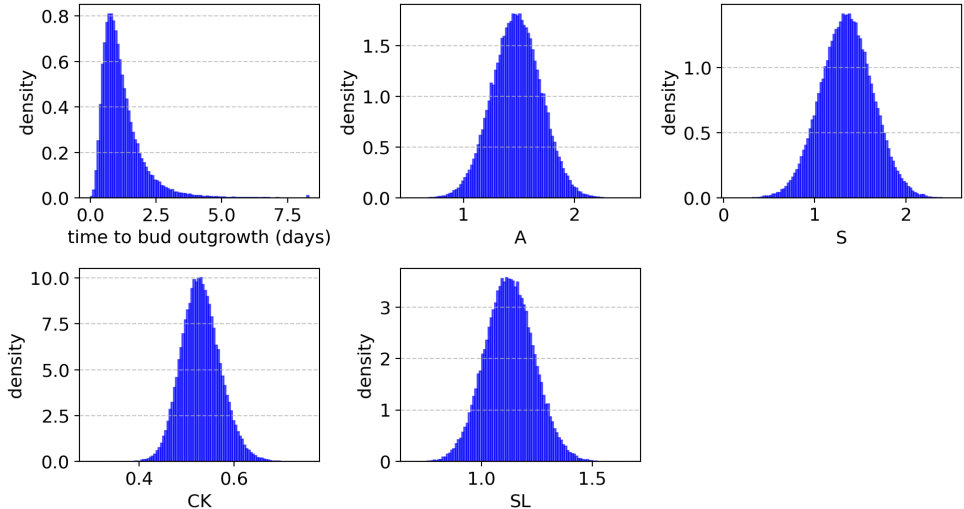

Figure 1: Distribution of the phenotype time to bud outgrowth $y_{\text{BO}}$, and intermediate traits auxin (A), sucrose (S), cytokinin (CK) and strigolactone (SL).

### 1.3 Model architecture

Each pathway subnetwork is a shallow fully-connected network (FCN) with hidden dimension $H = 32$, inter-layer sigmoid activations that aim to emulate the Hill-like dynamics of the shoot-branching network, and a one-dimensional output. The four pathway outputs are concatenated and passed through a final FCN with hidden dimension $F = 16$ to predict $y_{\text{BO}}$. Any SNPs not assigned to a pathway are fed into a residual FCN of the same size.

### 1.4 Experimental setup

All BINN experiments used the same training pipeline, implemented in PyTorch [6]. Below we summarize the key hyperparameters and procedures:

**Data splits and noise injection.** For each run, we randomly sample from the full set $N_{\text{train}}$ lines for training, $N_{\text{val}} = 0.1 \times N_{\text{total}}$ for validation, and $N_{\text{test}} = 0.2 \times N_{\text{total}}$ for testing. To test the robustness of model to noisy data we then add Gaussian noise to the targets:

$$y_i \leftarrow y_i \left(1 + \epsilon_i^{(y)}\right), \quad \epsilon_i^{(y)} \sim \mathcal{N}(0, \sigma_y^2),$$

and to each intermediate trait $p \in \{A, S, CK, SL\}$:

$$x_i^{(p)} \leftarrow x_i^{(p)} \left(1 + \epsilon_i^{(p)}\right), \quad \epsilon_i^{(p)} \sim \mathcal{N}(0, \sigma_{\text{inter}}^2),$$

where $\sigma_y = 10\%$ and $\sigma_{\text{inter}} = 5\%$.

**Optimization.** We trained each BINN with:

- **Optimizer:** Adam, initial learning rate $\alpha = 10^{-3}$, no weight decay.

4

- **Batch size:** 64.

- **Scheduler:** `ReduceLROnPlateau` on the validation loss, factor 0.5, patience 20 epochs.

- **Early stopping:** terminate if validation loss does not improve for 20 consecutive epochs, up to a maximum of 500 epochs.

**Repetitions, dataset sizes, and random seeds.** We evaluated each BINN configuration over nine logarithmically spaced training set sizes (500 to 20,000 samples), combined with varying noise levels and intermediate-trait availability. For each of these dataset sizes, we performed $n = 5$ independent runs using different random seeds. Genotypes were drawn once per size, and each seed controlled the train/validation/test split (80/10/10), the network weight initialization, and the Gaussian noise added to both the final phenotype and intermediate traits.

## 1.5 Baseline Models

To benchmark the performance of our BINN architectures, we compared against two standard models trained on the same synthetic data using our own implementation:

**Ridge regression.** We fit a linear model

$$\hat{y} = Xw + b,$$

with $L^2$ regularization on the weights. Concretely, the model implements

$$\mathcal{L}(w, b) = \frac{1}{N} \sum_{i=1}^{N} (y_i - (x_i^\top w + b))^2 + \lambda \|w\|_2^2,$$

where $x_i \in \mathbb{R}^p$ is the genotype vector, $y_i$ the target, and $\lambda = 0.01$ the regularization coefficient. Training proceeds by minimizing this loss via AdamW with the same learning rate and early-stopping schedule as the BINNs.

**Fully-connected network (FCN).** As a representative "black-box" nonlinear model, we trained a small deep network with two hidden layers of size 64 and ReLU activations:

$$\begin{aligned} h^{(1)} &= \text{ReLU}(W^{(1)}x + b^{(1)}), \\ h^{(2)} &= \text{ReLU}(W^{(2)}h^{(1)} + b^{(2)}), \\ \hat{y} &= W^{(3)}h^{(2)} + b^{(3)}. \end{aligned}$$

The model parameters $\{W^{(k)}, b^{(k)}\}$ are learned by minimizing mean-squared error:

$$\mathcal{L} = \frac{1}{N} \sum_{i=1}^{N} (y_i - \hat{y}_i)^2,$$

using the same optimizer, learning rate, batch size, and early-stopping criteria as for the BINNs.

## 1.6 Evaluation Metric

To quantify predictive performance, we compute the mean squared error (MSE) on the held-out test set for each model, dataset size, and random seed. Specifically, for each dataset size $N_{\text{train}}$, noise configuration, and intermediate-trait fraction, we perform $n = 5$ independent runs yielding test errors

$$\text{MSE}_r = \frac{1}{N_{\text{test}}} \sum_{i=1}^{N_{\text{test}}} \left(y_i - \hat{y}_i^{(r)}\right)^2, \quad r = 1, \ldots, n.$$

We then summarize these 5 values via their mean and standard deviation in the main text, and visualize their full distribution using boxplots.

# 2 Transcriptomics-derived BINN

## 2.1 Introduction

Linking genetic variation to phenotypic traits remains a central challenge in plant genomics. Torres-Rodriguez et al. [10] conducted a transcriptome-wide association study (TWAS) on the Wisconsin Diversity Panel, consisting 693 temperate adapted maize inbred lines from seven heterotic pools: stiff stalk (SS), non-stiff stalk (NSS), iodent (IDT), tropical, popcorn, sweet corn, and other/unknown, to identify genes controlling flowering time. The authors collected full-length RNA-seq within a tight two-hour window and recorded days to anthesis and silking at two Corn Belt locations, Nebraska and Michigan. From these data, they identified 21 genes exceeding stringent false-discovery thresholds and mapped both *cis*- and *trans*-eQTL, including loci in well-characterized flowering regulators. The study's power derives from its large sample size, precise sampling protocol, and deep sequencing depth, making it one of the most comprehensive TWAS datasets in maize.

Because this cohort provides the complete "genotype–expression–phenotype" triangle, it is ideal for evaluating our methodology. We hypothesize that embedding expression-derived sparsity into the network architecture improves predictive accuracy over genotype-only models, while still requiring only genotype data at inference. We constructed the BINN's sparsity mask through a two-step procedure. First, we trained an expression(biological domain knowledge)-to-phenotype (B2P) model using Elastic Net regression on the RNA-seq data for each flowering trait, tuning the regularization parameters via cross-validation. We then selected genes with non-zero coefficients. Second, we performed eQTL mapping on those candidate genes to identify their top SNP markers. The resulting SNP-gene pairs defined the binary mask connectivity matrix used to define our sparse genotype-to-expression-to-phenotype BINN architecture.

## 2.2 Dataset and Preprocessing

We use the real-world maize data provided in Torres-Rodriguez et al. 2024 [10], which includes flowering times (days to silking and anthesis) measured at two field locations (Nebraska and Michigan), RNA-seq gene expression data from leaf samples, and genotypes from

15.8 million single nucleotide polymorphism (SNP) markers. The dataset consists of 690 lines from the Wisconsin Diversity Panel, covering 7 heterotic groups (stiff stalk, non-stiff stalk, iodent, tropical, popcorn, sweet corn, and other/unknown) and expression values of 39,756 genes. We followed the data pre-processing steps described in Torres-Rodriguez et al. [10] and used the spatially-corrected flowering time values that were provided. Fig. 2 shows the distribution of the 4 phenotypes across all the 690 lines. We also dropped the unexpressed genes, with median TPM less than 0.01, from further analysis. This reduced the total number of genes to 24,874. Furthermore, markers with a minor allele frequency (MAF) below 5% were dropped, and PLINK [9] was used to prune highly correlated markers (`--indep-pairwise` 1000 500 0.5), leaving 3.2 million markers for further analysis. The pre-processed expression and marker data is, then, used for feature selection for the BINN model.

### 2.2.1 Feature Selection

The BINN architecture is sparsified by selecting a subset of genes and markers through a biologically-informed process. The genes are selected using a linear elastic net model and markers are selected through eQTL process. This reduces the input features (markers) by an order of 2 and the length of intermediate layer (genes) by an order of 1.

1. **Gene selection :** To identify a reduced set of significant genes influencing flowering time, we developed ElasticNet models utilizing expression data from 24,874 genes as predictors, with days to silking and anthesis for NE and MI (four distinct models) as the response variables. The expression data, measured in Transcripts Per Million (TPM), was normalized to a range of 0 to 2, following the methodology outlined by Torres-Rodriguez et al. [10]. The ElasticNet models were implemented using Scikit-Learn's ElasticNetCV[4], and we selected the number of genes based by varying the values of the L1 ratio by subsetting the features with non-zero coefficients. We used 5-fold cross-validation for model building. To assess the quality of the selected genes, we compared the overlap between genes identified as influencing flowering time in Torres-Rodriguez et al. [10] and those selected by the ElasticNet models. Table 1 presents the number of genes identified for various values of the L1 ratio, $(l)$, and the corresponding percentage of overlap with the flowering time genes reported in Torres-Rodriguez et al. [10], based on models trained using expression values from all the 690 lines. This analysis demonstrates that as the L1 ratio decreases, the number of selected genes increases, and a lower L1 ratio facilitates the identification of a greater number of significant genes.

2. **Marker selection :** For each gene selected through the ElasticNet model, significant markers were identified using eQTL analysis [2] as outlined in Torres-Rodriguez et al. [10], using mixed linear model [12] in rMVP [11] and including three principal components as covariates. For all the identified genes, the top 20 statistically significant markers are selected for training the BINN models.

Table 1: Assessment of the quality of the set of genes obtained using the ElasticNet models.

| Phenotype | # Significant flowering time genes from [10] | # genes from ElasticNet models | | | % overlap of significant genes | | |
|---|---|---|---|---|---|---|---|
| | | $l = 1.0$ | $l = 0.5$ | $l = 0.2$ | $l = 1.0$ | $l = 0.5$ | $l = 0.2$ |
| Days to Silking, MI | 17 | 317 | 568 | 970 | 70% | 82% | 100% |
| Days to Anthesis, MI | 16 | 320 | 412 | 816 | 69% | 81% | 94% |
| Days to Silking, NE | 16 | 234 | 413 | 756 | 69% | 81% | 100% |
| Days to Anthesis, NE | 14 | 259 | 383 | 760 | 64% | 100% | 100% |

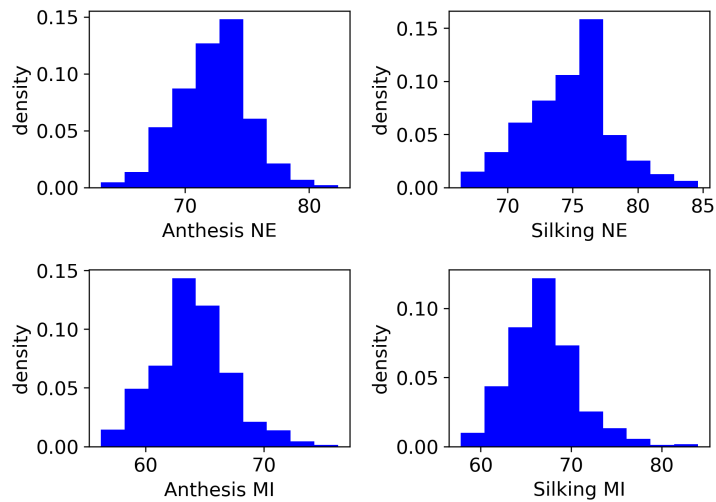

Figure 2: Distribution of the four phenotypes; days to anthesis and silking in NE and MI, for the real-world maize TWAS dataset

## 2.3 Model architecture

The BINN architecture comprises of three layers: an input layer consisting of the reduced set of markers, an intermediate layer in which each node represents a gene, and a final output layer that generates the predicted phenotype values. Every node (gene) in the intermediate layer is connected to 20 nodes (significant markers) from the input layer via a pathway sub-network, which is structured as a fully connected network (FCN) with a single hidden layer of dimension, $H = 128$, and a sigmoid activation function. The outputs from all the pathway sub-networks are concatenated and subsequently fed into another FCN, with a single hidden layer of dimension, $F = 128$, to predict the flowering time, $y_{\text{FT}}$.

## 2.4 Experimental setup

The experiments were designed to evaluate the performance of the BINN under conditions of sparse data. Consequently, all models were trained on 20% of the dataset while being tested on the remaining 80%. The entire dataset was divided into five non-overlapping

subsets, while ensuring that the lines from all heterotic groups were represented in same proportion as the full dataset. The exact number of train, validation and test lines used in each independent fold is shown in Table 2. To prevent any data leakage during feature selection, gene and marker selection were performed independently for each subset, ensuring that the corresponding features were utilized exclusively when training the model on that specific subset of data. For gene selection, we used an L1 ratio of $l = 0.1$ for ElasticNet models, with a 5-fold cross-validation. Table 3 presents the number of genes selected per phenotype for all the splits of data. For every selected gene, we selected 20 markers based on the eQTL analysis. It is important to note that expression data was used solely for feature selection to introduce sparsity in the network architecture and was not utilized for model training. Therefore, only genotype data is required for training and validation of the model.

Table 2: Number of train, validation and test lines for each subpopulation used in each fold (sparse-data scenario).

| Population | Train | Validation | Test |
|---|---|---|---|
| All | 65 | 65 | 560 |
| Others | 29 | 29 | 241 |
| SS | 15 | 15 | 129 |
| NSS | 9 | 9 | 76 |
| IDT | 5 | 5 | 47 |
| Tropical | 3 | 3 | 26 |
| Popcorn | 2 | 2 | 18 |
| Sweet corn | 2 | 2 | 23 |

We trained all the BINN models with Adam optimizer with an initial learning rate, $\alpha = 10^{-3}$, and batch size of 16. Training was terminated if the validation loss did not improve for 20 consecutive epochs, with a maximum limit of 500 epochs. Additionally, we used 5-fold cross-validation (CV) for training on the five subsets of data. As BINN is a neural network, it necessitates a validation dataset during training. However, the baseline models utilize the entire 20% of the data for training without requiring a validation set. To ensure a fair comparison between the models, we initially trained the BINN models using 16% of the data for training and 4% for early stopping/validation for each of the five CV folds. We recorded the epoch that yielded the best model performance for each of these runs and subsequently retrained the BINN model using 20% of the data for the same number of epochs without a validation set.

## 2.5 Baseline Models

We compare the performance of the BINN models against benchmark genotype-to-phenotype (G2P) and domain expression-to-phenotype (B2P) models, all of which are linear in nature. For G2P, we trained GBLUP and ridge regression models, whereas, for B2P, a ridge regression model was trained. All the ridge regression models were implemented using Scikit-Learn's [4] RidgeCV function with 3-fold cross-validation. We used the GBLUP implementation from BGLR [7] with 5000 iterations after 1000 burn-in iterations.

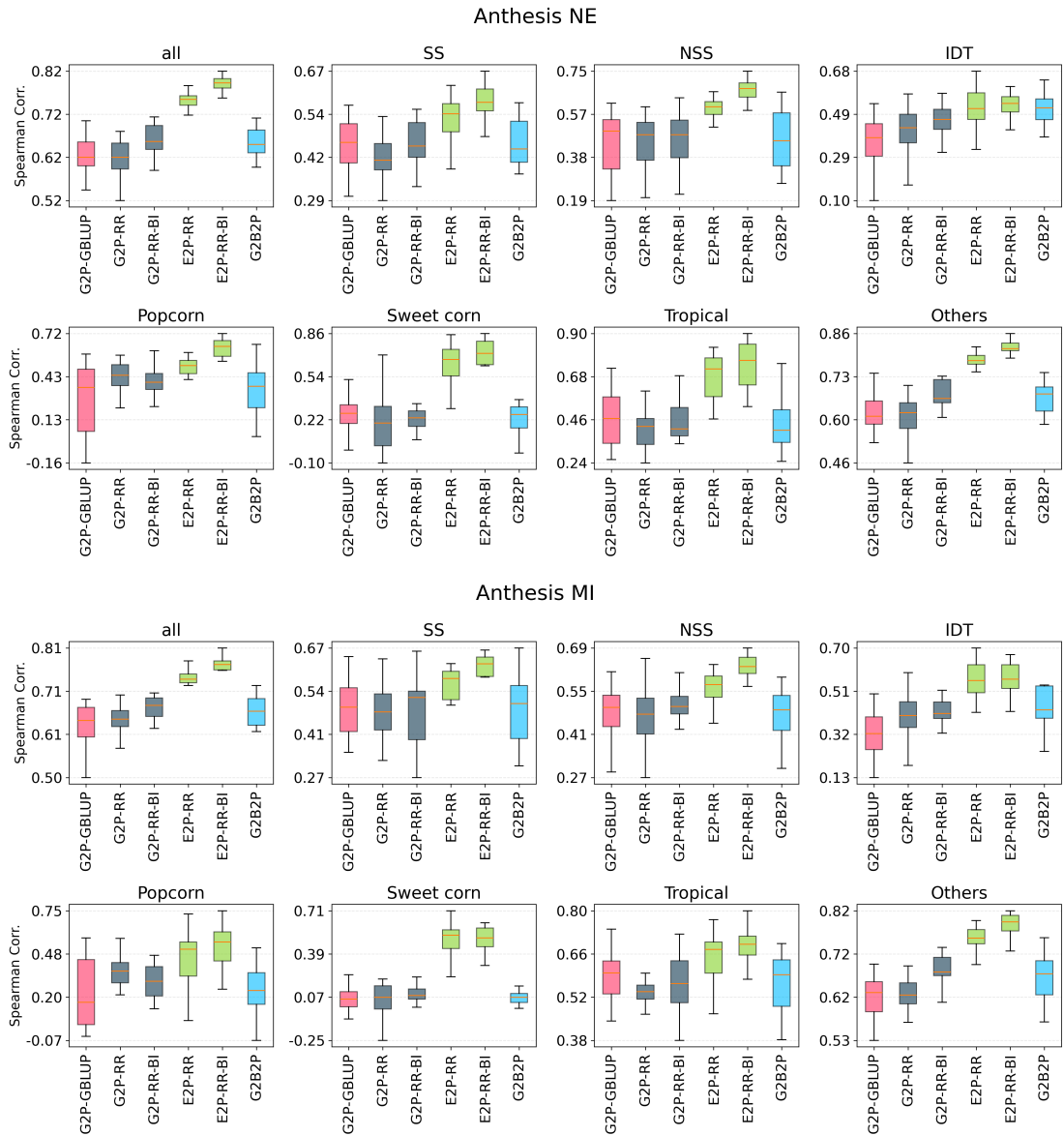

Figure 3: Test Spearman correlation distributions for prediction of days to Anthesis in NE and MI — comparing all the models (G2P on genotype, pink; B2P: Elastic Net on expression, green; G2B2P: BINN with network sparsity defined by lasso selected genes and eQTL-selected markers, blue) across five independent 20%/80% train/test splits, each with five-fold cross-validation. Results of three G2P models are shown - GBLUP (G2P-GBLUP), Ridge Regression on every 100th marker (G2P-RR) and Ridge Regression on bio-informed (eQTL-selected) markers (G2P-RR-BI). Additionally, results of two B2P models are also shown - Ridge Regression on all the genes (B2P-RR) and Ridge Regression on bio-informed (ElasticNet-selected) genes (B2P-RR-BI). Each subplot shows results for all lines pooled and for each of the seven distinct subpopulations.

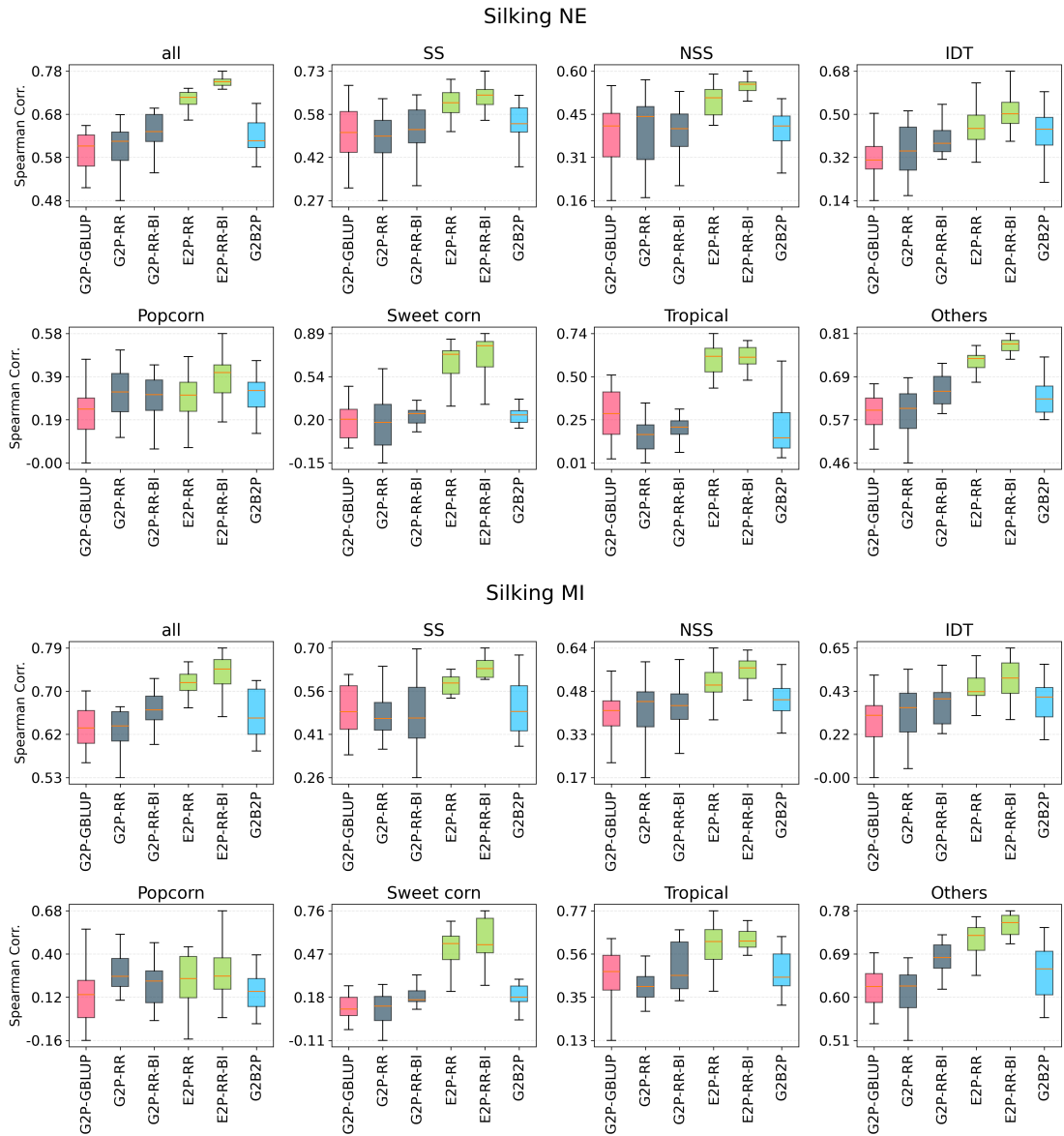

Figure 4: Test Spearman correlation distributions for prediction of days to Silking in NE and MI — comparing all the models (G2P on genotype, pink; B2P: Elastic Net on expression, green; G2B2P: BINN with network sparsity defined by lasso selected genes and eQTL-selected markers, blue) across five independent 20%/80% train/test splits, each with five-fold cross-validation. Results of three G2P models are shown - GBLUP (G2P-GBLUP), Ridge Regression on every 100th marker (G2P-RR) and Ridge Regression on bio-informed (eQTL-selected) markers (G2P-RR-BI). Additionally, results of two B2P models are also shown - Ridge Regression on all the genes (B2P-RR) and Ridge Regression on bio-informed (ElasticNet-selected) genes (B2P-RR-BI). Each subplot shows results for all lines pooled and for each of the seven distinct subpopulations.

Table 3: Number of genes selected per phenotype for different subset of data using $l = 0.1$.

| Data subset | Days to silking, MI | Days to anthesis, MI | Days to silking, NE | Days to anthesis, NE |
| --- | --- | --- | --- | --- |
| Split 1 | 1,271 | 1,041 | 1,010 | 933 |
| Split 2 | 1,230 | 995 | 1,071 | 912 |
| Split 3 | 1,079 | 919 | 969 | 941 |
| Split 4 | 1,188 | 985 | 1,044 | 848 |
| Split 5 | 1,235 | 1,018 | 1,042 | 933 |

The G2P models were trained using every 100th marker, as we found no significant difference in performance compared to using all 3.2 million markers. On the other hand, B2P model was trained on the expression (in TPM) of all 24,874 genes. Prior to model training, the TPM values were transformed using the Box-Cox normalization with a parameter $\lambda = 0.0282$ to approximate a normal distribution.

## 2.6 Evaluation Strategy

The performance of the BINN models are compared with that of the baseline models using Spearman's rank correlation coefficients [5]. Specifically, we calculated the correlation between the predicted and observed phenotype values, evaluated on the remaining 80% of the held-out data. Correlation values are computed for all five folds in each subset of data, as well as for each heterotic group represented in the dataset. It is important to note that since the correlation metric is computed for each fold using the fixed 80% held-out lines within a subset/split, this may underestimate the true model variance, potentially resulting in less variability in our test metrics. Additionally, we treated each heterotic group as an independent trial, aggregating Spearman correlations across the five folds, five data splits, and all four phenotypes, and then applied a paired t-test [3] to determine whether the BINN model's performance is significantly better than that of the G2P baseline. In the results reported in the main text we obtained $p = 2.23 \times 10^{-6}$, well below the 0.05 significance threshold, allowing us to reject the null hypothesis and conclude that BINN significantly outperforms the baseline model.

## 2.7 Additional Evaluation Results

In this section, we discuss additional experiments conducted to further evaluate the impact of biologically informed feature selection and network sparsity in BINNs. First, we compare the performance of the BINN model against various baseline models. Specifically, we consider three G2P baseline models - GBLUP, Ridge Regression using every 100th marker and Ridge Regression using eQTL-selected markers and two B2P models - Ridge Regression using all the genes and Ridge Regression using ElasticNet-selected genes. Figures 3 and 4 illustrate the performance of these models in predicting four flowering time traits — days to anthesis and silking in both NE and MI.

Furthermore, we conducted an ablation study to evaluate the impact of biologically informed network sparsity on the performance of the BINN model. This involved incorporating

12

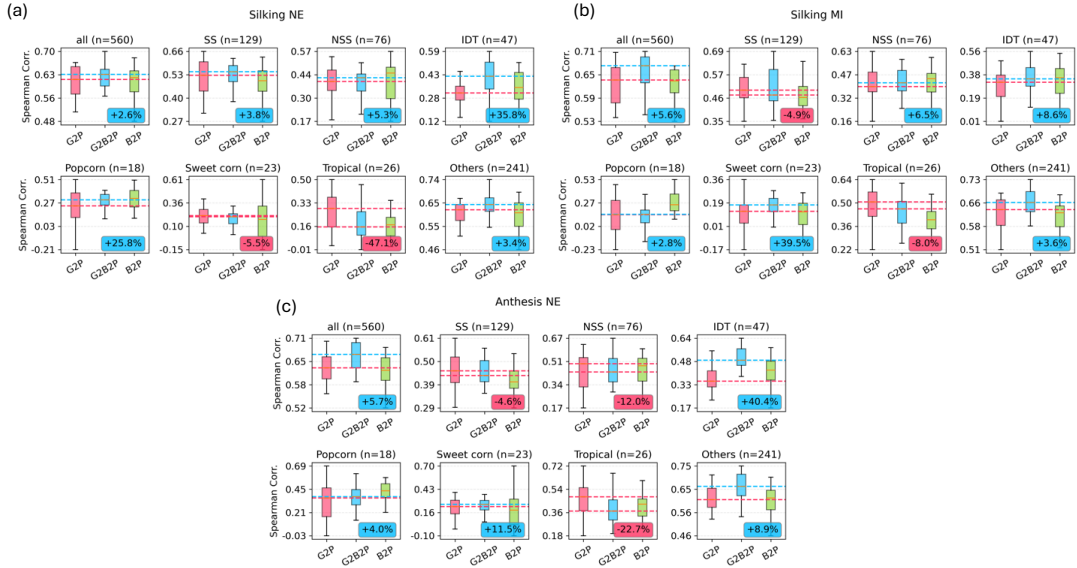

Figure 5: Test Spearman correlation distributions for (a) Silking NE, (b) Silking MI and (c) Anthesis NE - comparing three predictive models (G2P: GBLUP on genotype, pink; B2P: Elastic Net on expression, green; G2B2P: BINN with expression-informed sparsity, blue) across five independent 20%/80% train/test splits, each with five-fold cross-validation. Each subplot shows results for all lines pooled ("all") and for each of the seven distinct subpopulations (SS, NSS, IDT, Popcorn, Sweet corn, Tropical and Others). The pink dashed line marks the median of the G2P baseline; the dashed line for G2B2P is colored green when its median exceeds that baseline or red when it falls below. Legends report the median percent change of G2B2P relative to G2P.

randomly selected network sparsity into the BINN architecture and comparing its predictive performance with that of the baseline models. Figure. 7 presents the results for the BINN model, where network sparsity is defined by randomly selected genes and eQTL-selected markers. To ensure a fair evaluation, we maintained an equal number of randomly selected genes as in the biologically informed case. Figure. 8 displays similar results for network sparsity defined by randomly selected genes and markers.

Our observations from these analyses indicate that as the level of random connections in the network architecture increases—from randomly selected genes to randomly selected genes and markers—the performance of the BINN model significantly declines. Moreover, biologically informed markers and genes enhance the performance of the baseline models, as evidenced in Figures 3 and 4. These results underscore our finding that biologically informed sparsity imparts essential inductive bias to the model, thereby, improving its predictive accuracy.

## 3 Alternative BINN Architectures

In the main text, we described our canonical BINN design and its instantiation for (1) the synthetic shoot-branching problem and (2) the maize TWAS dataset. The following examples

13

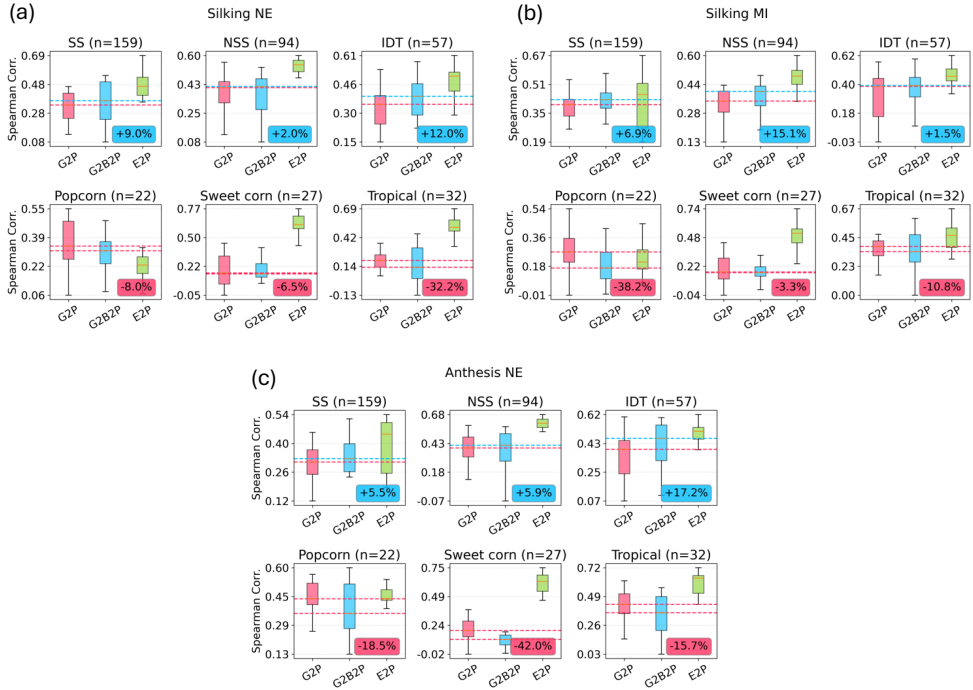

Figure 6: Test Spearman correlation distributions for leave-one-population-out experiments for (a) Silking NE, (b) Silking MI and (c) Anthesis NE - comparing three predictive models (G2P: Ridge Regression on genotype, pink; B2P: Elastic Net on expression, green; G2B2P: BINN with expression-informed sparsity, blue) across five independent 20% train splits, each with five-fold cross-validation. Each subplot shows results for the predictions on the six distinct subpopulations (SS, NSS, IDT, Popcorn, Sweet corn, and Tropical) using the models that are trained by **leaving out** the corresponding subpopulation from the training data. The pink dashed line marks the median of the G2P baseline; the dashed line for G2B2P is colored green when its median exceeds that baseline or red when it falls below. Legends report the median percent change of G2B2P relative to G2P.

are not exhaustive alternatives but illustrate the flexibility of our architecture in handling diverse multi-omics contexts:

### 3.1 Single-layer BINN

This model - used for the TWAS problem - has one intermediate omics layer (gene expression) feeding directly into the final integrator. It is appropriate when a single data type dominates predictive power, as in transcriptome-wide association studies without the availability of additional -omics data such metabolomics or proteomics.

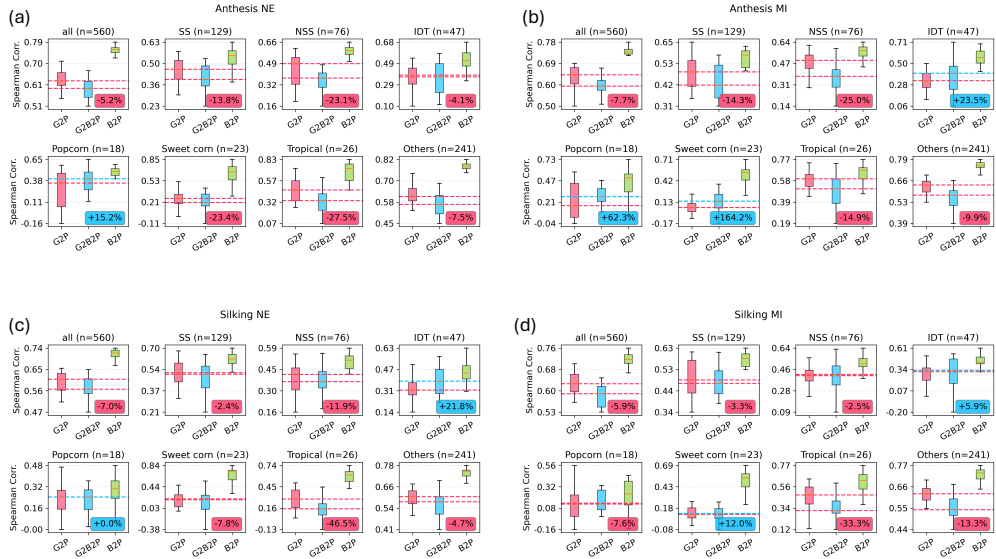

Figure 7: Test Spearman correlation distributions for four flowering-time traits—(a) Anthesis NE, (b) Anthesis MI, (c) Silking NE, and (d) Silking MI—comparing three predictive models. Here, the network sparsity of G2B2P (BINN) model is defined by **randomly selected genes** and **eQTL-selected markers**, blue). Models were trained across five independent 20%/80% train/test splits, each with five-fold cross-validation. Each subplot shows results for all lines pooled and for each of the seven distinct subpopulations.

## 3.2 Staggered-layer BINN

In cases with a single omics modality but with rich intra-layer dependencies, we allow pathway subnetworks to exchange information via sparse cross-connections. For example, in the shoot-branching network, auxin (A) and sucrose (S) subnetworks both feed into the cytokinin (CK) and strigolactone (SL) modules. These staggered links capture non-hierarchical, within-layer interactions, such as gene–gene or metabolite–metabolite crosstalk, before passing all signals to the final integrator.

## 3.3 Stacked-layer BINN

When multiple omics layers form a biological cascade, such as genotype → transcriptome → metabolome, we embed each layer as a separate set of subnetworks connected in series. The latent outputs of the first layer become the inputs to the second layer's pathway modules, and so on, before final integration. This design mirrors natural hierarchies (e.g. expression driving metabolite synthesis) and is appropriate whenever upstream molecular changes are known to mediate downstream effects.

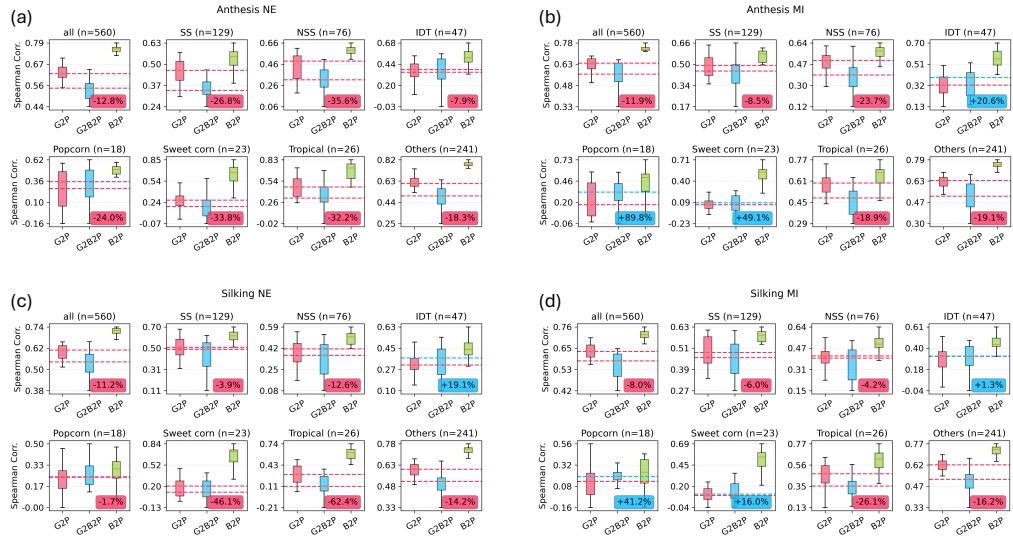

Figure 8: Test Spearman correlation distributions for four flowering-time traits—(a) Anthesis NE, (b) Anthesis MI, (c) Silking NE, and (d) Silking MI—comparing three predictive models. Here, the network sparsity of G2B2P (BINN) model is defined by **randomly selected genes** and **randomly selected markers**, blue). Models were trained across five independent 20%/80% train/test splits, each with five-fold cross-validation. Each subplot shows results for all lines pooled and for each of the seven distinct subpopulations.

### 3.4 Parallel-layer BINN

Each omics type (e.g. transcriptomics, proteomics, methylation) is processed by an independent subnetwork, and their latent outputs are concatenated only at the final integrator. This "late fusion" design suits orthogonal assays with minimal direct cross-talk, such as combining methylation and ATAC-seq to predict complex agronomic traits.

All variants share the same core elements, biological sparsity masks, shallow FCNs per module, and a unified integrator, yet differ in how modules interconnect, allowing users to tailor the model to their study's data structure and biological assumptions.

# References

[1] Jessica Bertheloot, François Barbier, Frédéric Boudon, Maria Dolores Perez-Garcia, Thomas Péron, Sylvie Citerne, Elizabeth Dun, Christine Beveridge, Christophe Godin, and Soulaiman Sakr. Sugar availability suppresses the auxin-induced strigolactone pathway to promote bud outgrowth. *New Phytologist*, 225(2):866–879, 2020.

[2] Beth Holloway, Stanley Luck, Mary Beatty, J-Antoni Rafalski, and Bailin Li. Genome-wide expression quantitative trait loci (eqtl) analysis in maize. *BMC genomics*, 12:1–14, 2011.

16

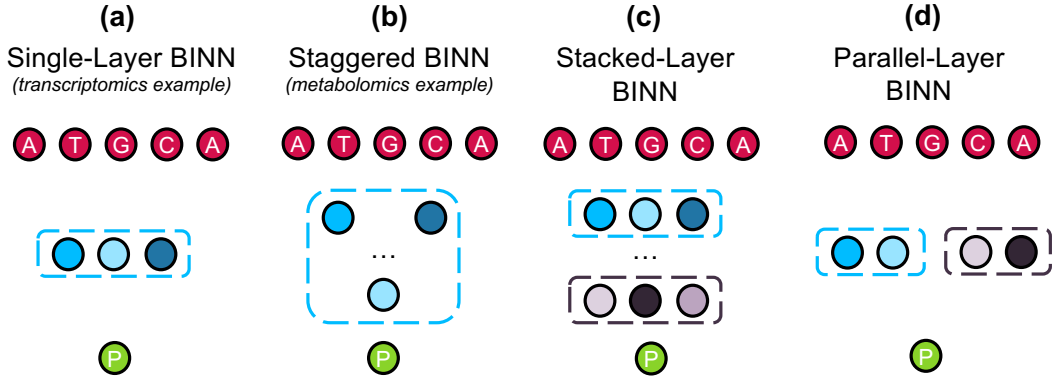

Figure 9: Flexible BINN architectures for diverse multi-omics scenarios. (a) Single-layer BINN: A single intermediate omics layer (e.g. gene expression) feeds directly into the final integrator; this configuration was used for the TWAS problem. (b) Staggered-layer BINN: Within a single omics modality, pathway subnetworks exchange information via sparse cross-connections. For example, in the shoot-branching network, both auxin (A) and sucrose (S) modules feed into the cytokinin (CK) and strigolactone (SL) subnetworks before final integration.(c) Stacked-layer BINN: Multiple omics layers are embedded in series - outputs from the first layer's subnetworks become inputs to the second layer's modules - mirroring cascades such as transcriptome → metabolome. (d) Parallel-layer BINN: Independent subnetworks process each omics type (e.g. transcriptomics, proteomics, methylation) in parallel, with their latent outputs concatenated only at the final integrator ("late fusion").

[3] Henry Hsu and Peter A Lachenbruch. Paired t test. *Wiley StatsRef: statistics reference online*, 2014.

[4] Oliver Kramer and Oliver Kramer. Scikit-learn. *Machine learning for evolution strategies*, pages 45–53, 2016.

[5] Leann Myers and Maria J Sirois. Spearman correlation coefficients, differences between. *Encyclopedia of statistical sciences*, 12, 2004.

[6] Adam Paszke, Sam Gross, Francisco Massa, Adam Lerer, James Bradbury, Gregory Chanan, Trevor Killeen, Zeming Lin, Natalia Gimelshein, Luca Antiga, Alban Desmaison, Andreas Kopf, Edward Yang, Zachary DeVito, Martin Raison, Alykhan Tejani, Sasank Chilamkurthy, Benoit Steiner, Lu Fang, Junjie Bai, and Soumith Chintala. Pytorch: An imperative style, high-performance deep learning library. *Advances in Neural Information Processing Systems*, 32, 2019.

[7] Paulino Pérez and Gustavo de Los Campos. Genome-wide regression and prediction with the bglr statistical package. *Genetics*, 198(2):483–495, 2014.

[8] Owen M Powell, Francois Barbier, Kai P Voss-Fels, Christine Beveridge, and Mark Cooper. Investigations into the emergent properties of gene-to-phenotype networks across cycles of selection: a case study of shoot branching in plants. *in silico Plants*, 4(1):diac006, 2022.

[9] Shaun Purcell, Benjamin Neale, Kathe Todd-Brown, Lori Thomas, Manuel AR Ferreira, David Bender, Julian Maller, Pamela Sklar, Paul IW De Bakker, Mark J Daly, et al. Plink: a tool set for whole-genome association and population-based linkage analyses. *The American journal of human genetics*, 81(3):559–575, 2007.

[10] J Vladimir Torres-Rodríguez, Delin Li, Jonathan Turkus, Linsey Newton, Jensina Davis, Lina Lopez-Corona, Waqar Ali, Guangchao Sun, Ravi V Mural, Marcin W Grzybowski, et al. Population-level gene expression can repeatedly link genes to functions in maize. *The Plant Journal*, 119(2):844–860, 2024.

[11] Lilin Yin, Haohao Zhang, Zhenshuang Tang, Jingya Xu, Dong Yin, Zhiwu Zhang, Xiaohui Yuan, Mengjin Zhu, Shuhong Zhao, Xinyun Li, et al. rmvp: a memory-efficient, visualization-enhanced, and parallel-accelerated tool for genome-wide association study. *Genomics, proteomics & bioinformatics*, 19(4):619–628, 2021.

[12] Zhiwu Zhang, Elhan Ersoz, Chao-Qiang Lai, Rory J Todhunter, Hemant K Tiwari, Michael A Gore, Peter J Bradbury, Jianming Yu, Donna K Arnett, Jose M Ordovas, et al. Mixed linear model approach adapted for genome-wide association studies. *Nature genetics*, 42(4):355–360, 2010.



\end{document}